\documentclass{article}

\usepackage{microtype}
\usepackage{graphicx}
\usepackage{subfigure}
\usepackage{booktabs} 

\usepackage{hyperref}

\usepackage[accepted]{icml2025}

\usepackage{amsmath}
\usepackage{amssymb}
\usepackage{mathtools}
\usepackage{amsthm}
\usepackage{amsmath}
\usepackage{amssymb}
\usepackage{multirow}
\usepackage{algorithm}
\usepackage{algcompatible}
\usepackage{mathtools}
\usepackage{booktabs}
\PassOptionsToPackage{dvipsnames}{xcolor}
\usepackage{soul}

\usepackage{wrapfig}
\DeclareMathOperator*{\argmax}{arg\,max}

\usepackage{colortbl}

\definecolor{lightgreen}{rgb}{0,0.9,0}
\definecolor{lightmintbg}{rgb}{.69,.86,.99}
\definecolor{lightmintgreen}{rgb}{0.8,0.98,0.81}
\definecolor{lightyellow}{rgb}{.99,.99,.85}
\definecolor{Grey}{gray}{0.5}
\usepackage{multirow}
\usepackage{caption}
\usepackage{pifont} 
\usepackage{booktabs} 
\usepackage{bbm}

\newcommand{\methodname}[0]{NeaR}
\newcommand{\abs}[1]{\lvert #1 \rvert}
\newcommand{\cmark}{\ding{51}} 
\newcommand{\xmark}{\ding{55}} 
\usepackage[capitalize,noabbrev]{cleveref}

\theoremstyle{plain}

\theoremstyle{definition}

\theoremstyle{remark}

\icmltitlerunning{NeaR}

\begin{document}

\twocolumn[
\icmltitle{Efficient Vocabulary-Free Fine-Grained Visual Recognition in the Age of Multimodal LLMs}




\begin{icmlauthorlist}
\icmlauthor{Hari Chandana Kuchibhotla}{iith}
\icmlauthor{Sai Srinivas Kancheti}{iith}
\icmlauthor{Abbavaram Gowtham Reddy}{cispa}
\icmlauthor{Vineeth N Balasubramanian}{iith}
\end{icmlauthorlist}

\icmlaffiliation{iith}{Indian Institute of Technology Hyderabad, India}
\icmlaffiliation{cispa}{CISPA Helmholtz Center for Information Security, Saarbrücken, Germany}

\icmlcorrespondingauthor{Hari Chandana Kuchibhotla}{ai20resch11006@iith.ac.in}

\icmlkeywords{Machine Learning, Multimodal LLMs}

\vskip 0.3in
]



\printAffiliations{}  

\begin{abstract}
Fine-grained Visual Recognition (FGVR) involves distinguishing between visually similar categories, which is inherently challenging due to subtle inter-class differences and the need for large, expert-annotated datasets. In domains like medical imaging, such curated datasets are unavailable due to issues like privacy concerns and high annotation costs. In such scenarios lacking labeled data, an FGVR model cannot rely on a predefined set of training labels, and hence has an unconstrained output space for predictions. We refer to this task as Vocabulary-Free FGVR (VF-FGVR), where a model must predict labels from an unconstrained output space without prior label information. While recent Multimodal Large Language Models (MLLMs) show potential for VF-FGVR, querying these models for each test input is impractical because of high costs and prohibitive inference times. To address these limitations, we introduce \textbf{Nea}rest-Neighbor Label \textbf{R}efinement (\methodname{}), a novel approach that fine-tunes a downstream CLIP model using labels generated by an MLLM. Our approach constructs a weakly supervised dataset from a small, unlabeled training set, leveraging MLLMs for label generation. \methodname{} is designed to handle the noise, stochasticity, and open-endedness inherent in labels generated by MLLMs, and establishes a new benchmark for efficient VF-FGVR.
\end{abstract}    
\vspace{-18pt}
\section{Introduction}
\label{sec_intro}
\vspace{-3pt}
Fine-Grained Visual Recognition (FGVR) is a task in computer vision that focuses on distinguishing between highly similar categories within a broader class~\cite{wei2021fine}. This task has gained increased importance in recent years given the success of foundation models on coarse-grained classification tasks. Traditional image classification aims to differentiate between dogs, cats, and birds, while FGVR aims to distinguish between different subcategories, such as \textit{Tennessee Warbler, Yellow-rumped Warbler} and \textit{Orange-crowned Warbler} among birds as an example. Fine-grained understanding is essential for a wide range of applications including biodiversity studies~\cite{liu2024novel,yang2020fine}, medical diagnosis~\cite{ridzuan2022self,nosubclass}, manufacturing~\cite{yang2022sensors}, fashion retail~\cite{cheng2022surveyfashion} and agriculture~\cite{yang2020fine}. 
\vspace{-12pt}
\begin{table}[t]
        \centering
        \scalebox{0.68}{
        \begin{tabular}{l|c|c|c|c}
        \toprule
        \textbf{Method} & \textbf{Training Time}& \textbf{Inference Time} & \textbf{Cost (\$)} & \textbf{cACC}\\
            \midrule
            FineR~\cite{finer} &$10\ h$&$1.12\ h$&$0$&$57.0$\\
            \midrule
             \textsuperscript{\normalsize \textdagger} GPT-4o & - & $17.5\ h$ & $\sim 100$ & $59.2$ \\
            ZS-CLIP + GPT-4o & - &$0.03$\ h&$1$&$54.6$\\
            \rowcolor{lightmintgreen}NeaR + GPT-4o (Ours) &$1.57\ h$&$0.03$\ h&$1$&$\mathbf{67.6}$\\
            \midrule
             \textsuperscript{\normalsize \ddag} LLaMA&-&$9.12\ h$&$0$&$48.4$\\
            ZS-CLIP + LLaMA& - & $0.03\ h$ & $0$ & $60.5$ \\
            \rowcolor{lightmintgreen}NeaR + LLaMA (Ours) & $1.57\ h$ & $0.03\ h$ & $0$ & $\mathbf{65.0}$\\
            \bottomrule
        \end{tabular}
        }
        \vspace{-5pt}
        \captionof{table}{\footnotesize Performance and cost metrics for different methods on benchmark FGVR datasets (computed over $32,503$ images from the test sets). 
        \textdagger\ = proprietary models, \ddag\ = open-source models (both used only for inference). Our method \methodname{} achieves a clustering accuracy (cACC described in \S~\ref{sec:experiments}) that exceeds even direct MLLM queries, at a fraction of cost and time taken.}
        \label{tab:costtable}
        \vspace{-14pt}
\end{table}

FGVR poses significant challenges due to the subtle differences between categories and the need for large annotated datasets. Typically, fine-grained classification datasets are annotated by domain experts who meticulously examine each image and assign a corresponding label. 
However, when such domain experts are unavailable, not only is labeled data unavailable, but the underlying fine-grained categories of interest may also be unknown. We refer to this task as Vocabulary-Free FGVR (VF-FGVR), where the vocabulary of fine-grained labels is not provided. In this context, pre-trained Multimodal Large Language Models (MLLMs), which are trained on vast corpora of image-text data and possess extensive world knowledge, provide a contemporary solution~\cite{reid2024gemini}. MLLMs excel at zero-shot multimodal tasks such as Visual Question Answering (VQA). By recasting FGVR as a VQA problem, where we ask the question {\footnotesize \texttt{`What is the best fine-grained class label for this image?'}}, MLLMs can predict fine-grained labels in a zero-shot manner without prior training on a specific dataset. The ability of MLLMs to operate in a VF setting presents a promising approach for domains where curated, labeled datasets are scarce or unavailable.
However, querying such large models for each test input is computationally expensive and time-consuming, making their usage impractical at scale. As shown in Table~\ref{tab:costtable}, performing inference with GPT-4o~\citep{achiam2023gpt} on $32,203$ test images from benchmark FGVR datasets takes $\approx 17.5$ hours of querying and incurs a cost of $\approx$ USD \$100. 

\textit{Hence, especially considering the need for sustainable AI systems, there is an imminent need for an efficient VF-FGVR system that performs on par with MLLMs on the fine-grained understanding task, while being efficient in terms of time and computational resources required.} We focus on this problem in this work.
To this end, we only consider access to a small unlabeled set of training images belonging to the classes of interest, with no information regarding the individual class names or even the total number of classes. Such unlabeled datasets are relatively easy to obtain for many domains -- for instance, a collection of unlabeled photos of exotic birds. We empirically show in \S~\ref{sec:experiments} that having about $3$ images per class is sufficient to learn an FGVR system that can perform inference for any number of test samples in the considered experiments. To solve this VF-FGVR task, we propose to label this training set by querying an MLLM for each image. Such a noisily labeled training set can be used in different ways -- a simple strategy would be to utilize this MLLM supervision to build a zero-shot classifier over the set of generated labels using a pre-trained CLIP~\cite{Radford2021LearningTV} model. Another way would be to naively fine-tune the CLIP model using the generated labels. In both these approaches, test images can now be classified by the CLIP model without the need for expensive forward passes through an MLLM. Although such simple methods are efficient in compute and time taken, they fall short on performance as shown in \S~\ref{sec:experiments}. This is because MLLM outputs are inherently noisy and open-ended, so the generated labels do not necessarily provide strong supervision. 

To address these limitations, we propose \textbf{Nea}rest-Neighbor Label \textbf{R}efinement (\textbf{NeaR}), a method designed to learn using the noisy labels generated by an MLLM. Our approach first constructs a candidate label set for each image using the generated labels of other similar images. In line with prior work on learning with noisy labels (LNL)~\cite{dividemix}, we partition the dataset into clean and noisy samples. We then design a label refinement scheme for both partitions that can effectively combine information from the constructed candidate set and the generated label. Finally, to address the open-ended nature of MLLM outputs, we incorporate a label filtering mechanism to truncate the label space. Our method \methodname{} thus enables us to handle the inherently noisy and open-ended labels generated by MLLMs, allowing us to effectively fine-tune a downstream CLIP model. As shown in Table~\ref{tab:costtable}, for GPT-4o, our approach can achieve performance exceeding that of direct inference while incurring only $1/100^{th}$ of the total inference cost, and requiring a negligible fraction of inference time.

Our key contributions can be summarized as: \textbf{(i)} To the best of our knowledge, this is the first work that uses state-of-the-art MLLMs to build a cost-efficient vocabulary-free fine-grained visual recognition system, \textbf{(ii)} We propose a pipeline that can handle noisy and open-ended labels generated by an MLLM. Our proposed method \methodname{} leverages similarity information to construct a candidate label set for each image which is used to mitigate the impact of label noise. We also design a label filtering mechanism to improve classification performance. \textbf{(iii)} We perform a comprehensive set of experiments showing that \methodname{} outperforms existing works and the MLLM-based baselines we introduce for VF-FGVR, achieving this in a cost-efficient way.

\section{Related Work}

\noindent\textbf{Fine-Grained Visual Recognition.} FGVR~\cite{WahCUB_200_2011, Maji2013FineGrainedVC} aims to identify sub categories of an object, such as various bird species, aircraft type etc. FGVR has been extensively studied in prior work~\cite{wei2021fine}. 
A key limitation of these methods is their reliance on annotated datasets, which are often unavailable in many important domains like e-commerce and medical data. With advancements in Vision-Language Models and MLLMs, the burden of dataset annotation can be alleviated, reducing the need for extensive human effort.
\textbf{Foundation Models for VF-FGVR.} 
Recent advancements in MLLMs have led to models demonstrating strong zero-shot performance across a wide range of multimodal tasks \cite{li2023blip, achiam2023gpt, reid2024gemini, llama, liu2023improvedllava}. These MLLMs can be applied to VF-FGVR by framing the task as a VQA problem. MLLMs are broadly categorized into two types: (1) Proprietary models, such as GPT-4o \cite{achiam2023gpt} and GeminiPro \cite{reid2024gemini}, and (2) Open-source models, including BLIP-v2 \cite{li2023blip}, LLaVA-1.5-7B~\cite{liu2023improvedllava}, LLaMA-3.2-11B \cite{llama} and Qwen2-7B~\cite{Qwen2VL}.
As shown in Table~\ref{tab:costtable}, for both types of MLLMs, performing inference for every test point remains computationally expensive and time-consuming. To address this, recent works have developed more efficient solutions for VF-FGVR. For instance, \cite{finer, liu2024rar, cased} propose pipelines that use cascades of MLLMs. FineR~\cite{finer} presents a pipeline combining VQA systems, Large Language Models (LLMs), and a downstream CLIP model, leveraging unsupervised data to build a multimodal classifier for inference. RAR \cite{liu2024rar} uses a multimodal retriever with external memory, retrieving and ranking top-k samples using an LLM. CaSED \cite{cased} approaches VF-FGVR by accessing an external database to retrieve relevant text for a given image. Nevertheless, these methods are often complex and do not fully exploit the advancements in MLLMs, resulting in suboptimal performance.
\textbf{Prompt Tuning.} Prompt-tuning methods add a small number of learnable tokens to the input while keeping the pretrained parameters unchanged. The tokens are fine-tuned to enhance the performance of large pre-trained models on specific tasks. Context Optimization (CoOp)~\cite{coop} was the first to introduce text-based prompt tuning, replacing manually designed prompts like \textit{"a photo of a"} with adaptive soft prompts. We study the impact of our method under other prompt-tuning methods such as VPT~\cite{Jia2022VisualPT} and IVLP~\cite{ivlp} in Appendix\S~\ref{app:prompting_strategies}.
\textbf{Learning with Noisy Labels.} ~\cite{Arpit2017ACL} demonstrated the memorization effect of deep networks, showing that models tend to learn clean patterns before fitting noisy labels. To mitigate this,~\cite{Han2018CoteachingRT, pmlr-v97-chen19g} introduce iterative learning methods to filter out noisy samples during training. ~\cite{pmlr-v97-arazo19a} proposed a mixture model-based approach to partition datasets into clean and noisy subsets, leading to more reliable training. Building on these insights, DivideMix~\cite{dividemix}, a state-of-the-art LNL method, combines semi-supervised learning with data partitioning to achieve superior performance on noisy datasets. JoAPR~\cite{guo2024joapr} is a contemporary approach to fine-tune CLIP on noisy few-shot data. We compare against JoAPR in \S~\ref{tab:joapr_result_small}.
\vspace{-2mm}
\begin{figure*}[t]
    \centering
    \includegraphics[width=0.9\linewidth]{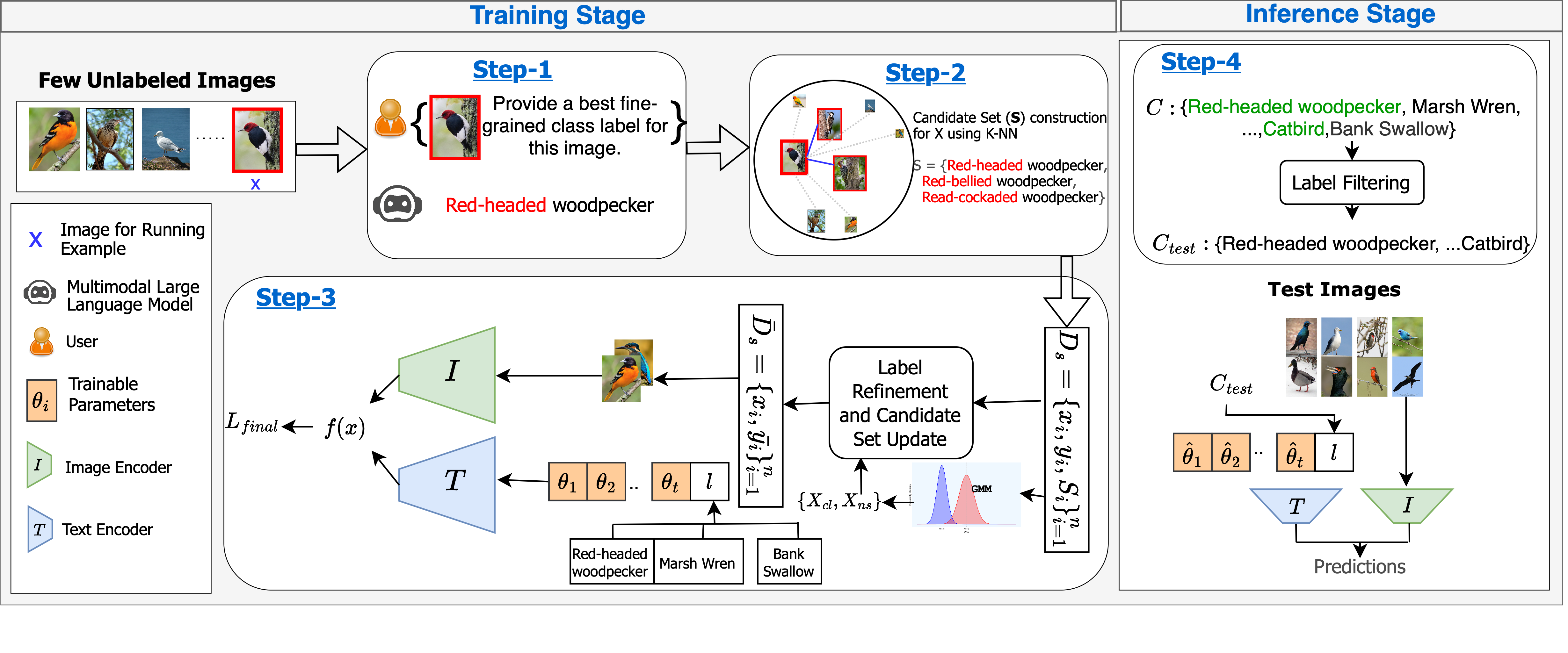}
    \vspace{-9pt}
    \caption{\footnotesize Overview of our proposed method, \textbf{NeaR}, for Vocabulary-Free Fine-Grained Visual Recognition (VF-FGVR). In the Training Stage, we start with a few unlabeled images. Step-1: An MLLM generates a best-estimate fine-grained label (e.g., "Red-headed woodpecker") for each image. Step-2: A candidate label set is constructed using K-Nearest Neighbors, capturing related fine-grained classes. Step-3: The model is fine-tuned using a CLIP-based architecture. A GMM is applied to the loss to partition the data into clean and noisy samples. Based on this split, a label refinement mechanism is used to further update and refine the labels. The final loss, $L_{final}$, is then computed, and the model parameters are updated accordingly. In the Inference Stage, we apply label filtering to limit the label space (Step-4). Our approach handles the noise and open-ended nature of MLLM-generated labels, significantly reducing inference time and cost while maintaining performance.}
    \vspace{-9pt}
    \label{fig:mainfigure}
\end{figure*}
\vspace{-2pt}
\section{Methodology}
\label{sec:methodology}
\vspace{-2pt}

As shown in Table~\ref{tab:costtable}, although MLLMs are capable of performing VF-FGVR, labeling every test image is expensive and time consuming which limits their practical application. To address these limitations, we propose \textbf{Nea}rest-Neighbor Label \textbf{R}efinement (\textbf{\methodname{}}), a method designed to leverage MLLMs efficiently for VF-FGVR. Our approach begins by constructing a candidate label set for each unlabeled training image as described in \S~\ref{sec:methodology_2}. Next we partition the data into `clean' and `noisy' samples using a Gaussian Mixture Model (GMM) and the small-loss rule~\cite{Arpit2017ACL}. In \S~\ref{sec:methodology_3}, we show how to refine labels of noisy samples by incorporating information from the candidate sets. The final loss function is a simple cross-entropy loss, where the refined labels serve as the targets. As outlined in \S~\ref{sec:preliminaries}, using MLLM outputs directly can result in an excessively large label space, which can hinder performance. To address this, \methodname{} incorporates a label filtering mechanism to truncate the label space, which boosts classification performance while improving efficiency. Once the CLIP model is fine-tuned, it can classify test images without requiring additional expensive forward passes through an MLLM, significantly improving inference efficiency. Our methodology \methodname{} is able to exceed the performance of direct MLLM-based classification at just a fraction of the cost and compute. An overview of our methodology is presented in Figure~\ref{fig:mainfigure}, and the pseudocode is detailed in Algorithm\S~\ref{mainalgo} in the appendix. We begin by discussing the necessary preliminaries. 
\vspace{-4mm}
\subsection{Preliminaries}
\label{sec:preliminaries}
\noindent\textbf{Problem Formalization.} 
We consider a setting where only a small, unlabeled training set of $n$ images $X=\{x_i\}_{i=1}^n$, $x_i\in\mathcal{X}$ is available. We assume that this training set has at least $m$-shot samples for each class of the unknown ground-truth class name set $\mathcal{G}$. We also study a more realistic scenario where there is class imbalance in \S~\ref{sec:ablation_study} and observe that the performance of \methodname{} does not degrade. Note that no further information about $\mathcal{G}$ is known, including its cardinality. For each image $x_i$, we obtain a class name $l_i=L(x_i,p)$ from an MLLM $L$, where $p$ is a simple text prompt \texttt{`Provide a best fine-grained class name for this image.'} that guides the MLLM to generate a fine-grained class name for the image. The generated dataset $\mathcal{D}=\{(x_i,l_i)\}_{i=1}^n$ consists of $n$ image-label pairs, and the output space of class names is denoted by $\mathcal{C}=\bigcup_{i=1}^n l_i$. W.l.o.g. we assume $\mathcal{C}$ is lexicographically ordered and we denote the $k=\abs{\mathcal{C}}$ labels by $\mathcal{C}=\{c_1,c_2,\dots,c_k\}$. Let $y_i\in\{0,1\}^{k}$, $i\in[n]$, denote the one-hot encoding of the text label $l_i$ for image $x_i$ i.e $y_{i}^j=1$ if $l_i=c_j$ and $0$ otherwise. Due to the inherent noise and stochastic nature of MLLM generated labels, it is common to have $\lvert \mathcal{C}\rvert > \lvert\mathcal{G} \rvert$. Furthermore, the number of generated labels increases with the size of the training set and can become prohibitively large, hampering training and reducing efficiency of the downstream CLIP classifier.

\noindent\textbf{CLIP Classifier:} CLIP~\cite{Radford2021LearningTV} consists of an image encoder $\mathcal{I}$ and a text encoder $\mathcal{T}$ trained contrastively on image-text pairs. For VF-FGVR, we first query an MLLM to build a few-shot dataset $\mathcal{D}$ with label space $\mathcal{C}$. CLIP classifies image $x$ as $\hat{l}(x)=\argmax_{l\in\mathcal{C}} cos(\mathcal{I}(x), \mathcal{T}(l))$. We call this baseline ZS-CLIP. Note that ZS-CLIP does not leverage the paired supervision in $\mathcal{D}$ and thus serves as a simple baseline. CLIP can benefit from fine-tuning on a small labeled datast. CoOp~\cite{coop} is a prompt-tuning approach that adds a small number of learnable tokens $\theta$ to the class name. We denote CLIP's class predictions by $f_{\theta}(x)\in\Delta^{k}$, where $\Delta^{k}$ denotes the $k-1$ simplex.
The prompts are trained on the dataset $\mathcal{D}$ by minimizing cross-entropy loss $\displaystyle L_{CE}(\theta)= \frac{-1}{n}\sum\limits_{i=1}^n \sum\limits_{j=1}^{k}y_{i}^{j}log(f_{\theta}^{j}(x_i))$. Naive finetuning using CoOp on a dataset generated by an MLLM can be susceptible to label noise. Our \methodname{} method mitigates this issue by refining labels through nearest-neighbor information. The following sections detail how \methodname{} constructs candidate sets, learns with these sets, and filters labels for effective downstream performance.
\subsection{\methodname{}: Candidate Set Construction}
\label{sec:methodology_2}
The label generated by an MLLM in response to a prompt may vary significantly from the ground truth class label for each image. We propose to leverage local geometry to mitigate the noise in generated labels.  More formally, we make the \textit{manifold assumption}, which suggests that similar images should share similar or identical class labels~\cite{Iscen2022LearningWN,Li2022NeighborhoodCE}. This is particularly useful when the label $l_i$ assigned to image $x_i$ by the MLLM is incorrect, which we refer to as a noisy label. By constructing a candidate label set, we increase the likelihood of including the true label or a semantically closer alternative in the candidate set rather than relying solely on the potentially incorrect label provided by the MLLM. Table~\ref{tab:candidate_evaluation} shows that the semantic similarity between the best label in the candidate set and the ground truth is higher than that of the noisy single label, supporting our hypothesis. We use CLIP's pretrained image encoder $\mathcal{I}$ to extract image features of the entire training set $X$. For each image $x_i$, we select the top-$\kappa$ most similar images (including $x_i$ itself) and gather their corresponding labels to form the candidate set $S_i = (l_i, l_1, \ldots, l_{\kappa-1})$. In this work, we choose $\kappa=3$. The resulting dataset is reconstructed as $\mathcal{D}_s = \{(x_i, l_i, S_i)\}_{i=1}^n$, incorporating the candidate sets rather than single labels alone. An alternative way of noise mitigation is to have the MLLM directly generate a candidate set of `top-$\kappa$' labels for each image, instead of just a single label. However this approach does not make use of similarity information between images, as each candidate set is now generated independently, leading to an excessively large label space. We empirically demonstrate the effectiveness of our nearest-neighbor based candidate set generation over other alternatives in Appendix \S~\ref{sec:appx_direct_candidates}.
\begin{table}
    \centering
    \scalebox{0.71}{
    \begin{tabular}{l|c|c|c|c|c}
    \toprule
        \textbf{Method} & \textbf{Bird-200} & \textbf{Car-196} & \textbf{Dog-120} &\textbf{Flower-102} &\textbf{Pet-37} \\
        \toprule
        MLLM Labels &\large $69.9$&\large$78.6$&\large$70.9$&\large$57.9$&\large$84.8$\\ 
        Random CS &\large$72.8$&\large$78.9$&\large$73.6$&\large$61.6$&\large$85.9$\\
        \rowcolor{lightmintgreen} K-NN  CS (ours) &\large $\mathbf{78.0}$&\large $\mathbf{81.3}$&\large $\mathbf{75.5}$&\large $\mathbf{65.8}$&\large $\mathbf{87.4}$\\
         \bottomrule
    \end{tabular}
    }
    \vspace{-5pt}
    \caption{\footnotesize The table compares the quality of labels generated by the MLLM, Random candidate set (CS), and K-NN CS using sACC. While Random CS modestly increases the likelihood of including the true label compared to using MLLM labels directly, K-NN CS significantly outperforms Random CS, generating a superior candidate set and validating our hypothesis.}
    \vspace{-0.5cm}
    \label{tab:candidate_evaluation}

\end{table}
\vspace{-2mm}
\subsection{\methodname{}: Learning With a Candidate Set}
\label{sec:methodology_3}
We treat the candidate sets as a source of supplementary similarity information to be used in conjunction with the label. As explained below, for a \textit{noisy} image, where the initial label $l_i$ is incorrect, we propose relying on the candidate set $S_i$ to mitigate the impact of noise. Conversely, for a \textit{clean} image, we can trust and utilize its generated label $l_i$.

\noindent\textbf{Detecting Noisy Samples.}
\label{sec:methodology_4}
It has been demonstrated in~\cite{Arpit2017ACL} that models tend to learn clean samples before noisy ones, resulting in lower loss values for clean samples. Following DivideMix~\cite{dividemix}, for every training epoch, we fit a two-component Gaussian Mixture Model (GMM) over the cross entropy loss values of all training samples $\{L(x_i,l_i)\}_{i=1}^n$, where $L(x_i,l_i)=-\sum_{j=1}^k y_{i}^{j}log(f_{\theta}^{j}(x_i))$, $y_i$ is the one-hot encoding of $l_i$. The component with the smaller mean value models the clean samples, while the other component models the noisy ones. The posterior probability $w_i=\mathbb{P}_{GMM}(clean|x_i)$ computed from the fitted GMM is used to model the likelihood that a sample $x_i$ is clean. This GMM is refitted for every training epoch, enabling dynamic estimation of label noise over time. We now partition the training data into clean $X_{cl}=\{x_i\in X \mid w_i\geq \tau\}$ and noisy $X_{ns}=X\setminus X_{cl}$ sets based on clean probability threshold $\tau$. We use the average clean posterior as an adaptive threshold for every epoch, i.e $\tau=\frac{1}{n}\sum\limits_{i=1}^n w_i$. The effect of different thresholding strategies is presented in Appendix \S~\ref{sec:appx_additional_results}.   

\noindent\textbf{Warm-up.} Warm-up strategies are commonly used to speedup convergence and stabilize training. As demonstrated in~\cite{dividemix}, an initial warm-up phase allows the model to learn the clean samples better,  resulting in better separation between the losses of clean and noisy samples. During warm-up, we train prompts for a few epochs (10 in our experiments) by minimizing the cross entropy loss over the generated labels $l_i$ with one-hot representation $y_i$:
\vspace{-1mm}
\begin{equation*}
    L_{warmup}(\theta) = \frac{-1}{n}\sum\limits_{i=1}^n \sum\limits_{j=1}^k y_{i}^{j}\cdot log(f_{\theta}^{j}(x_i))
\end{equation*}
This warm-up step lays the groundwork for effective training by allowing the model to initially focus on images labeled correctly by the MLLM.

\noindent\textbf{Candidate Set Guided Label Refinement.} Following the initial warm-up phase, we make a forward pass over the entire training set at each training epoch to fit a GMM and partition data into clean and noisy samples $X_{cl}$ \& $X_{ns}$ as described earlier. We model the confidence of the candidate set $S_i$ by a vector $q_i\in \mathbb{R}^{k}$ for each image $i\in[n]$, initialized as $q_{i}^j=\frac{1}{\abs{S_i}}$ if $c_j \in S_i$ and $0$ otherwise. This initialization reflects uniform confidence over classes belonging to the candidate set, and zero for non-members.

A candidate set is derived from neighboring images and provides a broader view of possibly correct labels. Our approach constructs refined labels to effectively leverage 
this additional information. We propose to construct refined labels for clean and noisy images differently. For an image $x_i$ with one-hot label $y_i$ and candidate set confidence $q_i$, we construct a refined label $\bar{y}_i$ as:
\vspace{-4pt}
\begin{align*}
\bar{y}_i = 
\begin{cases}
    \text{shrp}\big(w_i \cdot y_i + (1 - w_i) \cdot f_{\theta}(x_i), T\big), \text{\quad if } x_i\in X_{cl} \\
    \text{rsc}\big(\text{shrp}(w_i \cdot q_i + (1 - w_i) \cdot f_{\theta}(x_i), T), q_i\big), \text{ o/w}
\end{cases}
\end{align*}
where $w_i$ is the GMM clean posterior probability, and $f_{\theta}(x_i)$ denotes the CLIP model class probabilities with learnable prompts $\theta$. The sharpen function $shrp(y, T)^i = (y^i)^{\frac{1}{T}}/\sum\limits_{j=1}^k (y^j)^{\frac{1}{T}}$, as defined in~\cite{mixmatch}, adjusts a probability distribution $y$ to be more confident using a temperature $T$.
The rescale function $rsc(y, q)^i = (y\odot q)^i/\sum\limits_{j=1}^k (y\odot q)^j$ rescales a probability $y$ with the current confidence estimates of the candidate set $q$.
This ensures that the refined label has non-zero probabilities only for the candidate labels. For both clean \& noisy images, we update candidate set confidence to be used in the next epoch as $q_i = \text{rsc}(f_{\theta}(x_i), 1[q_i])$
where $1[q_i]$ is $1$ at non-zero indices. Prompts are learned by minimizing the cross-entropy loss between the refined labels $\bar{y}$ and CLIP model output.
\begin{equation*}
\vspace{-2mm}
    L_{final}(\theta) = \frac{-1}{n}\sum\limits_{i=1}^n\sum\limits_{j=1}^{k} \bar{y}_{i}^{j}\cdot log(f_{\theta}^{j}(x_i))
\end{equation*}
\noindent\textbf{Connection to PRODEN.} Our loss is similar in spirit to losses designed for Partial Label Learning (PLL), such as PRODEN~\cite{provablyPLL}, which allow learning when only candidate labels are present. However unlike the PLL setting, our candidate sets are constructed for every image using noisy MLLM outputs, and may not contain the true label. Furthermore, our method uniquely benefits from access to an initial `best-estimate' label $l_i$ generated by the MLLM, which is not exploited by traditional PLL algorithms. This best estimate label allows us to differentiate clean samples and helps training convergence by transferring knowledge from clean to noisy samples through iterative updates of $q$.

\subsection{\methodname{}: Label Filtering}
\vspace{-0.5mm}
Although we train on the entire label set $\mathcal{C}$, we observe that many labels are noisy and can be removed from the inference time label space. Let $F_{clip}=\{c_i\mid \exists x\in X \text{ s.t } i=\argmax\limits_{j\in[k]} f_{\theta}^{j}(x) \}$ be a filtered set of labels which are predicted by CLIP on the training set. Let $F_{cand}=\{c_i\mid \text{ s.t } i=\argmax\limits_{j\in[k]} q_i^j\}$ be another filtered set of labels which are predicted using just the candidate sets. We propose to keep only those labels which belong to both sets. The evaluation time label space is $\mathcal{C}_{test} = F_{clip} \cap F_{cand}$
and the inference time prediction of an image $x$ is $\hat{l}(x)=\argmax\limits_{l\in \mathcal{C}_{test}} sim(\mathcal{I}(x), \mathcal{T}_{\hat{\theta}}(l))$, where $\hat{\theta}$ are the learned prompts. Label filtering is effective as shown in Table \S~\ref{tab:label_filtering}.
\vspace{-5pt}
\section{Experiments and Results}
\label{sec:experiments}
\vspace{-5pt}

In this section, we comprehensively evaluate the classification performance of \methodname{} for the VF-FGVR task.
We begin by describing the datasets, metrics and benchmark methods we compare against.

\begin{table*}
    \centering
    \scalebox{0.77}{
    \begin{tabular}{l|cccccccccc|cc}
    \toprule
    \textbf{Method}&\multicolumn{2}{c}{\textbf{Bird-200}}&\multicolumn{2}{c}{\textbf{Car-196}}&\multicolumn{2}{c}{\textbf{Dog-120}}&\multicolumn{2}{c}{\textbf{Flower-102}}&\multicolumn{2}{c}{\textbf{Pet-37}}&\multicolumn{2}{|c}{\textbf{Average}}\\
    \cmidrule{2-13}
         & cACC & sACC & cACC & sACC & cACC & sACC & cACC & sACC & cACC & sACC & cACC & sACC\\
         \midrule
         \color{Grey}ZS-CLIP-GT (Upper Bound)&\color{Grey}57.2&\color{Grey}80.1&\color{Grey}64.0&\color{Grey}66.5&\color{Grey}60.2&\color{Grey}77.6&\color{Grey}70.8&\color{Grey}79.7&\color{Grey}87.5&\color{Grey}92.0&\color{Grey}68.0&\color{Grey}79.2\\
         \midrule
         CaSED &25.6 &50.1&26.9&41.4&38.0&55.9&67.2&52.3&60.9&63.6&43.7&52.6 \\
        FineR &51.1 &69.5&49.2&63.5&48.1&64.9&63.8&51.3&72.9&72.4&57.0&64.3 \\
        RAR&51.6&69.5&53.2&63.6&50.0&65.2&63.7&53.2&74.1&74.8&58.5&65.3\\
        \midrule
         \textsuperscript{\normalsize \textdagger}GPT-4o&68.8 &85.2&37.4&61.5&71.1&80.4&50.5&51.6&68.2&83.5&59.2&72.4 \\
         ZS-CLIP-GPT4o&48.8&72.5&42.9&59.5&43.8&69.1&18.2&53.0&68.2&78.7&54.6&66.6\\
         CoOp-GPT-4o&54.4&75.4&51.9&59.8&60.4&72.9&70.4&51.7&83.5&86.3&64.1&$\mathbf{69.2}$\\
         \rowcolor{lightmintgreen}NeaR-GPT4o&55.8&75.6&57.0&60.0&61.6&74.4&80.6&52.1&82.9&84.0&\cellcolor{lightmintgreen}$\mathbf{67.6\textbf{(+3.5\%)}}$&\cellcolor{lightmintgreen}$\mathbf{69.2}$\\
         \midrule
         \textsuperscript{\normalsize \textdagger}Gemini Pro&66.1 &82.7&35.4&62.8&65.8&81.2&45.3&54.3&71.3&85.7&56.8&73.3\\
         ZS-CLIP-GeminiPro&51.7&74.6&41.6&61.7&58.9&72.6&57.7&49.1&71.7&78.6&56.3&67.3\\
         CoOp-GeminiPro&55.2&75.9&50.2&61.5&62.7&73.8&68.3&51.2&81.6&83.8&63.6&69.2\\
         \rowcolor{lightmintgreen}NeaR-GeminiPro&55.9&76.0&54.9&61.1&64.7&75.4&77.9&53.2&79.4&80.8&\cellcolor{lightmintgreen}$\mathbf{66.6\textbf{(+3\%)}}$&\cellcolor{lightmintgreen}$\mathbf{69.3\textbf{(+0.1\%)}}$\\
         \midrule
        \textsuperscript{\normalsize \ddag}Qwen2-VL-7B-Instruct&53.0&75.3&45.6&63.7&69.7&78.8&84.8&72.7&77.7&85.1&66.2&75.1\\
        ZS-CLIP-Qwen2&41.0&66.0&50.8&60.8&59.3&70.5&66.7&55.2&72.4&77.2&58.0&65.9\\
        CoOp-Qwen2&51.0&72.1&52.1&61.9&62.5&73.4&77.0&65.0&83.4&87.5&65.2&$\mathbf{72.0}$\\
        \rowcolor{lightmintgreen}NeaR-Qwen2&48.9&72.0&55.6&63.2&62.0&73.3&81.4&68.0&84.6&86.8&\cellcolor{lightmintgreen}$\mathbf{66.5\textbf{(+1.3\%)}}$&71.7(-0.3\%)\\
        
        \midrule
        \textsuperscript{\normalsize \ddag}LLaMA-3.2-11B&41.4&70.6&14.4&61.6&55.0&71.8&66.0&63.6&65.1&82.0&48.4&69.9\\
        ZS-CLIP-LLaMA&48.7&66.3&45.8&60.6&57.4&65.9&74.8&58.4&76.0&78.4&60.5&65.9\\
        CoOp-LLaMA&49.2&68.7&45.5&60.7&58.4&68.4&75.9&59.8&74.4&79.2&60.7&67.4\\
        \rowcolor{lightmintgreen}NeaR-LLaMA&51.0&70.2&52.6&60.9&59.2&70.2&78.6&61.7&83.5&86.2&\cellcolor{lightmintgreen}$\mathbf{65.0{\textbf{(+4.3\%)}}}$&\cellcolor{lightmintgreen}$\mathbf{69.8\textbf{(+2.4\%)}}$\\

         \bottomrule
    \end{tabular}
    }
    \vspace{-5pt}
    \caption{\footnotesize ZS-Zero Shot, \textsuperscript{\large \textdagger} proprietary models used for inference, \textsuperscript{\large \ddag} open-source models used for inference. Our results shown here are for $\kappa=3$ and $m=3$. The first row is ZS-CLIP performance when the ground-truth label space is given, serving as an upper bound. The second partition consists of contemporary VF baselines of which FineR~\cite{finer} is best performing. We outperform FineR by a large margin, even when using weaker open-source MLLMs. The next four partitions are for labels generated by various MLLMs. We compare \methodname{} against CoOp within each partition, and highlight best numbers in \textbf{bold}. Our method \methodname{} outperforms all contemporary baselines, as well as ZS-CLIP and CoOp baselines for a variety of MLLMs.}
    \label{tab:maintable}
    \vspace{-5pt}
\end{table*}

\noindent \textbf{Datasets:}
We perform experiments on five benchmark fine-grained datasets:  CaltechUCSD Bird-200~\cite{WahCUB_200_2011}, Stanford Car-196~\cite{khosla2011novel},  Stanford Dog-120~\cite{krause20133d}, Flower-102~\cite{nilsback2008automated}, Oxford-IIIT Pet-37~\cite{parkhi2012cats}. Following~\cite{finer}, for each dataset, \methodname{} and other baselines only have access to $m$ unlabeled training images per class. Unless specified otherwise, we assume $m=3$. Results for $1\leq m\leq 10$ are shown in Figure~\ref{fig:diff_shots}.

\noindent\textbf{Baselines:} We compare our method \methodname{} against \underline{four} different classes of baseline methods. \textbf{(i)} \underline{Direct Inference on MLLMs.} For every test image, we directly query an MLLM for a fine-grained label using a text prompt such as \texttt{`What is the best fine-grained class name for this image?'}. We evaluate two proprietary MLLMs -- GPT-4o~\cite{achiam2023gpt} and GeminiPro~\cite{reid2024gemini} and two strong open-source MLLMs, LLaMA-3.2-11B-Vision-Instruct~\cite{llama} and Qwen2-VL-7B-Instruct~\cite{Qwen2VL}. In the Appendix \S~\ref{sec:appx_additional_results}, we show results on two other weaker open-source MLLMs, BLIP-2~\cite{li2023blip} and LLaVA-1.5~\cite{liu2023improvedllava}. \textbf{(ii)} \underline{Contemporary VF Baselines.} We consider three contemporary baselines which do not require expert annotations but use foundational models to perform VF-FGVR -- CaSED~\cite{cased}, FineR~\cite{finer} and RAR~\cite{liu2024rar}. \textbf{(iii)} \underline{ZS-CLIP with MLLM label space.}  As described in \S~\ref{sec:preliminaries}, we can perform zero-shot classification using pre-trained CLIP over the label space generated by querying various MLLMs on training images. We consider four variants of ZS-CLIP -- ZS-CLIP-GPT4o, ZS-CLIP-GeminiPro, ZS-CLIP-LLaMA, and ZS-CLIP-Qwen2. \textbf{(iv)} \underline{Prompt Tuning Baselines.} Following CoOp as described in \S~\ref{sec:preliminaries}, we directly perform prompt-tuning using the labels generated by an MLLM. We consider four variants -- CoOp-GPT4o, CoOp-GeminiPro, CoOp-LLaMA, and CoOp-Qwen2. 

\noindent \textbf{Evaluation Metrics:} In the VF-FGVR setting, \methodname{} as well as all other baselines operate in an unconstrained label space, making accuracy an invalid metric since the predicted labels may never exactly match the ground-truth labels. Following~\cite{finer, cased}, we evaluate performance using two complementary metrics: \textit{Clustering Accuracy} (cACC) and \textit{Semantic Accuracy} (sACC). cACC measures the ability of the model to group similar images together. For $M$ test images with ground-truth labels $y^{\star}$ and predicted labels $\hat{y}$, cACC is computed as $\max\limits_{p\in\mathcal{P}(\hat{\mathcal{Y}})} \frac{1}{M}\sum\limits_{i=1}^{M} \mathbbm{1}(y^{\star}_i = p(\hat{y}_i))$, where $\mathcal{P}(\hat{\mathcal{Y}})$ is the set of all permutations of the generated labels. 
Since cACC disregards the actual label name, it does not measure if the predictions are semantically correct. Despite this limitation, cACC is a strong evaluation metric and is widely used in areas such as GCD~\cite{vaze2022gcd}, where the goal is to assess consistency of predictions rather than exact label semantics. 
Semantic closeness is captured by sACC, which measures the cosine similarity between Sentence-BERT~\cite{bert} embeddings of the predicted and ground-truth labels. As observed in~\cite{finer}, sACC is a more forgiving metric than cACC, because embedding based similarity methods can capture general semantics even for completely distinct labels. We hence consider cACC as representative of the model's performance, with sACC acting as a sanity check to ensure that the predicted labels remain meaningful.

\noindent\textbf{Implementation Details:} We use CLIP ViT-B/16~\cite{Radford2021LearningTV} as the VLM, whose image encoder we also use to find the $\kappa$-nearest neighbors, with $\kappa=3$ by default. The default number of shots is $m=3$, and we use the few-shot training splits provided by FineR. For both the CoOp baseline and our method, we introduce 16 trainable context vectors. We use SGD as the optimizer and train for 50 epochs, with 10 warmup epochs. Further implementation details are provided in Appendix \S~\ref{sec:appx_implementation}.  

\vspace{-7pt}
\subsection{Main Results}
\label{sec:main_results}
\vspace{-5pt}
In this section we compare \methodname{} against baselines on five fine-grained datasets. In addition to the considered baselines, we benchmark against JoAPR~\cite{guo2024joapr}, a state-of-the-art noisy label learning method designed for CLIP, and against PRODEN~\cite{provablyPLL}, a widely used partial label learning algorithm. 

\noindent\textbf{Benchmarking \methodname{} Against Baseline Methods:} We evaluate \methodname{} against the four categories of baselines introduced in \S~\ref{sec:experiments} -- Direct MLLM inference, contemporary VF methods, zero-shot CLIP, and prompt-tuned CLIP. The results are shown in Table~\ref{tab:maintable}, with all numbers reported for $3$-shot training images. The first partition of the table, ZS-CLIP-GT, is the performance of pre-trained CLIP when provided with the ground-truth label space, serving as an upper bound. Notably, \methodname{}-GPT-4o achieves an average cACC just $\mathbf{-0.4\%}$ below this upper bound, demonstrating it's effectiveness. The next partition consists of contemporary methods that can perform VF-FGVR.  Out of these, FineR~\cite{finer} is conceptually closest to ours as it uses a combination of an LLM and a VQA system to construct a training-free CLIP based classifier. We outperform FineR on all datasets by a margin of at least $\mathbf{+8\%}$ in average cACC, even when using labels from open-source MLLMs. 
Moreover, as shown in Table~\ref{tab:costtable}, \methodname{} is significantly more efficient in terms of computation time.

The next four partitions in Table~\ref{tab:maintable} report results using labels generated by GPT-4o, GeminiPro, LLaMA-3.2 and Qwen2 respectively. Within each partition, we first present results for direct inference with the MLLM, followed by ZS-CLIP, CoOp, and finally \methodname{}. Across all MLLMs, \methodname{} performs the best on average cACC, showing gains of at least $\mathbf{+3\%}$ over the CoOp baseline for GPT-4o, GeminiPro and LLaMA-3.2, and a gain of $\mathbf{+1.3\%}$ over CoOp for Qwen2. Furthermore, we observe a large performance gain for the difficult Car-196 dataset, where \methodname{}-LLaMA shows a gain of $\mathbf{+7.1\%}$ in cACC over CoOp-LLaMA. These results highlight that \methodname{} effectively learns from the imperfect labels generated by MLLMs, leading to robust and efficient fine-grained classification.

\noindent \textbf{Comparison against PRODEN:}
Our loss function resembles those used in Partial Label Learning (PLL), such as PRODEN~\cite{provablyPLL}, which are designed to handle learning with only candidate sets. To study the efficacy of traditional PLL approaches, we replace the traditional cross-entropy loss used in CoOp with PRODEN, and learn prompts using the candidate sets directly. The results are shown in Table~\ref{tab:joapr_result_small}, where \methodname{} outperforms PRODEN by a large margin of $4.3\%$ in cACC and $3.2\%$ in sACC. Unlike in traditional PLL where candidate sets are assumed to include the correct label, our candidate sets are generated for each image using noisy MLLM outputs and may not always contain the true label. Also, \methodname{} uniquely benefits from an initial "best-estimate" label $l_i$ from the MLLM, which traditional PLL methods do not exploit. As described in \S~\ref{sec:methodology_3}, this best-estimate label is used to find "clean" images which have a higher probability of being correctly labeled. Knowledge from these clean samples helps resolve the ambiguity in candidate sets, improving performance.

\noindent\textbf{Comparison against JoAPR~\cite{guo2024joapr}, a Contemporary Noisy Label Learning Method for CLIP:}
\setlength{\columnsep}{4pt}
\setlength{\intextsep}{4pt}
\begin{wraptable}[11]{r}{0.3\textwidth}
    \centering
    \scalebox{0.9}{
    \begin{tabular}{l|cc}
    \toprule
    \textbf{Method} &\multicolumn{2}{c}{\textbf{Average}} \\
    \midrule
         & cACC & sACC \\
    \midrule
        JoAPR-LLaMA &60.4&68.9\\
        PRODEN & 60.6 & 66.6 \\
        \rowcolor{lightmintgreen}NeaR-LLaMA & \textbf{65.0} & \textbf{69.8} \\
    \bottomrule
    \end{tabular}
    }
    \caption{ \footnotesize NeaR outperforms JoAPR and PRODEN with LLaMA generated labels, showing an improvement of $+4.6\%$ and $+4.4\%$ in average cACC respectively.}
    \label{tab:joapr_result_small}
\end{wraptable}
JoAPR is a prompt-tuning method designed to fine-tune CLIP on noisy few-shot data. In Table~\ref{tab:joapr_result_small}, we show the results of using JoAPR to learn from noisy LLaMA generated labels. Our method \methodname{} outperforms JoAPR by $\mathbf{+4.6\%}$ in average cACC, and by $\mathbf{+0.9\%}$ in average sACC. For JoAPR we use the default configuration suggested in the paper. These gains highlight that generic noisy-label learning methods, which expect structured noise (such as flips) within a closed label set, do not fully address the challenges posed by open-ended MLLM outputs.  
By incorporating similarity information, performing candidate set guided label refinement, and performing label filtering, \methodname{} provides a robust solution to the VF-FGVR problem.

\vspace{-7pt}
\subsection{Ablation Studies}
\label{sec:ablation_study}
\vspace{-4pt}
In this section we study the impact of each module of \methodname{}. \methodname{} leverages common techniques from existing literature on learning with noisy labels. As described in \S~\ref{sec:methodology_3}, we have a warm-up phase followed by a partition step where we fit a GMM on loss values to partition the data into clean and noisy samples. These two steps are common to many methods that attempt to learn with noisy labels~\cite{pmlr-v97-arazo19a, dividemix, guo2024joapr}, and are not unique to our method. We now study the impact of our novel additions -- candidate set guided label refinement, and label filtering. The results of this study are shown in Tab~\ref{tab:ablation}.

\noindent \textbf{Impact of Candidate Set.} The candidate label set for an image is used to refine labels of noisy samples. As described in \S~\ref{sec:methodology_4}, it is also used to aid in label filtering. To study the effect of removal of the candidate set, we refine the labels of the noisy samples as $\bar{y}_i=\text{shrp}(f_{\theta}(x_i), T)$, i.e we only used the sharpened CLIP pseudolabel. We also remove the candidate set based filtering $F_{cand}$, as defined in \S~\ref{sec:methodology_4}. Row 3 of Table~\ref{tab:ablation} gives the result of removal of the candidate set. We observe a drop of $\mathbf{-1.6\%}$ in cACC of the flowers-102 dataset, compared to our \methodname{} result in Row 4 where all components are present. 

\begin{table} 
\centering
\scalebox{0.8}{ 
\begin{tabular}{cccc|ccc}
\toprule 
Warmup & GMM & Candidate Set& Label filtering& cACC & sACC \\
\midrule
\cmark & \cmark & \xmark & \xmark & \large $73.9$ & \large $60.2$ \\
\cmark & \cmark & \cmark & \xmark & \large $75.6$ & \large $60.6$ \\
\cmark & \cmark & \xmark & \cmark &\large $76.9$ & \large $60.1$ \\
\rowcolor{lightmintgreen}\cmark & \cmark & \cmark & \cmark &\large $\mathbf{78.6}$&\large $\mathbf{61.7}$ \\
\bottomrule 
\end{tabular} 
} 
\caption{\footnotesize Ablation study of NeaR demonstrating the effectiveness of our novel components candidate set and label filtering. We report the ablations on Flowers-102 dataset.}
\label{tab:ablation}
\vspace{-15pt}
\end{table}

\noindent \textbf{Impact of label filtering.} The results of Rows 1 \& 2 of Table~\ref{tab:ablation} indicate the importance of performing label filtering. The results in Row 1 act as a simple LNL baseline to our method. We observe a drop of $\mathbf{-4.6\%}$ cACC and $\mathbf{-1\%}$ sACC for the flowers dataset, from our method in Row 4. This ablation clearly indicates that our components are crucial to learn from noisy MLLM labels. The result of Row 2 studies the impact of removing label filtering in isolation. We observe a drop of $\mathbf{-2.9\%}$ cACC compared to our method \methodname{} in Row 4. These ablations confirm that candidate-set guided label refinement and label filtering are integral for performance.

\noindent \textbf{Imbalanced Training Data:} We study the realistic scenario of class imbalance in the few-shot training data.
\setlength{\columnsep}{5pt}
\setlength{\intextsep}{5pt}
\begin{wraptable}[13]{r}{0.3\textwidth}
    \centering
    \scalebox{0.65}{
    \begin{tabular}{l|cc}
    \toprule
    \textbf{Method} & \multicolumn{2}{c}{\textbf{Average}} \\
    \midrule
         &cACC & sACC \\
         \midrule
        FineR&51.9&61.2\\
        ZS-CLIP-LLaMA &59.7&64.5\\
        CoOp-LLaMA&60.0&66.5\\
        \rowcolor{lightmintgreen}NeaR-LLaMA     & \textbf{65.7 {(+5.7\%)}} & \textbf{70.3{ (+3.8\%)}}\\
         \bottomrule
    \end{tabular}
    }
    \caption{\footnotesize Performance for long-tail class distribution of training data, where head classes have $4 \leq m \leq 10$ samples and tail classes have $m=3$ samples. Both \methodname{} and CoOp retain performance on imbalanced data.\methodname{} outperforms CoOp by $+5.7\%$ in average cAcc. }
    \label{tab:long_tail_smaller}
\end{wraptable}
We simulate a long-tail distribution where we randomly select a small number of head classes (10 classes for pet-37) which have $4\leq m\leq 10$ samples, and the remaining tail classes have $m=3$ samples. We show results in Table~\ref{tab:long_tail_smaller}. We observe that there is no degradation in performance for both CoOp and \methodname{}. Infact, we note slightly better cACC and sACC values for \methodname{} on account of the slight increase in the training data. The expanded table is in Appendix \S~\ref{sec:appx_additional_results}. 

\begin{figure}[ht]
    \centering
    \includegraphics[width=1.0\linewidth]{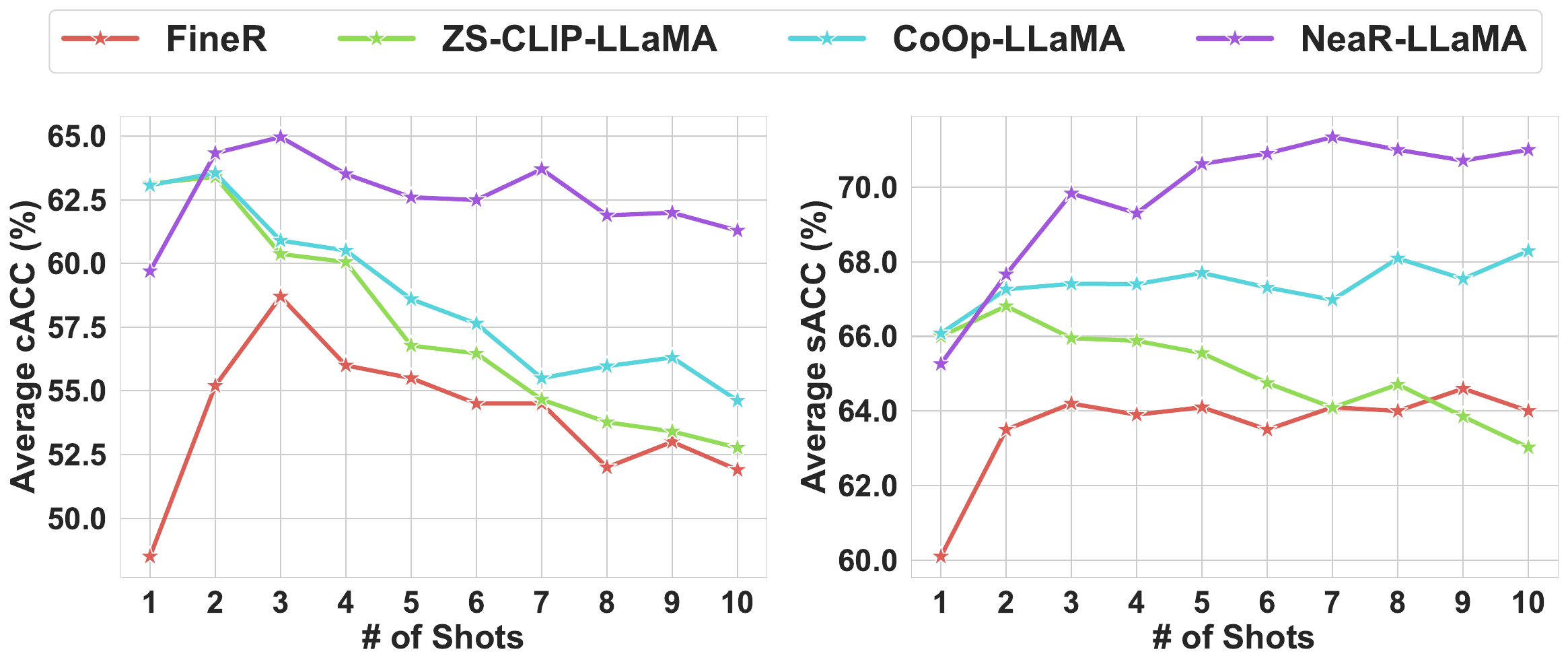}
    \vspace{-17pt}
    \caption{\footnotesize Effect of varying $m$, number of images per class in training data for labels generated by LLaMA. \methodname{} (in purple) outperforms CoOp (blue) for all $m\geq 2$ in average cACC \& sACC.}
    \label{fig:diff_shots}
    \vspace{-5pt}
\end{figure}

\noindent \textbf{Analysis on Number of Shots $\mathbf{m}$ in Training Data:} 
We explore the effect of the number of images used per class, as presented in Figure~\ref{fig:diff_shots}.
We consistently use $\kappa=3$ for candidate set construction across all shots. For $m=1$, our method performs poorly due to excessive label filtering. However, as the number of shots increases, our candidate set is more informative and performance improves markedly. Our proposed method, NeaR, outperforms CoOp for all $m\geq 2$, especially at higher shots where there are more noisy labels. We also observe that cACC drops with increasing $m$ due to increase in the size of the test time label space.

\noindent\textbf{Performance of \methodname{} on Different CLIP Backbones:} 
\begin{wraptable}[11]{r}{0.3\textwidth}
    \centering
    \scalebox{0.6}{
    \begin{tabular}{c|l|c|c}
    \toprule
    & \textbf{Method} & \textbf{Avg. cACC} & \textbf{Avg. sACC} \\
    \midrule
    \multirow{2}{*}{\rotatebox{90}{\parbox{0.75cm}{\centering \footnotesize RN50}}} 
    & CoOp-LLaMA & 18.6 & 45.0 \\
     &\cellcolor{lightmintgreen} NeaR-LLaMA &\cellcolor{lightmintgreen}\textbf{22.7 (+4.1\%)} & \cellcolor{lightmintgreen}\textbf{49.8 (+4.8\%)} \\
    \midrule
    \multirow{2}{*}{\rotatebox{90}{\parbox{0.85cm}{\centering \footnotesize RN101}}} 
    & CoOp-LLaMA & 21.8 & 45.6 \\
    & \cellcolor{lightmintgreen}NeaR-LLaMA &\cellcolor{lightmintgreen} \textbf{23.6}\textbf{ (+1.8\%)} &\cellcolor{lightmintgreen}\textbf{50.5 (+4.9\%)} \\
    \midrule
    \multirow{2}{*}{\rotatebox{90}{\parbox{0.9cm}{\centering \footnotesize ViT-B/16}}} 
    & CoOp-LLaMA & 60.7 & 67.4 \\
    &\cellcolor{lightmintgreen} NeaR-LLaMA &\cellcolor{lightmintgreen}\textbf{65.0 (+4.3\%)} &\cellcolor{lightmintgreen} \textbf{69.8 (+2.4\%)} \\
    \midrule
    \multirow{2}{*}{\rotatebox{90}{\parbox{0.9cm}{\centering \footnotesize ViT-B/32}}} 
    & CoOp-LLaMA & 55.7 & 66.2 \\
    &\cellcolor{lightmintgreen} NeaR-LLaMA &\cellcolor{lightmintgreen} \textbf{61.1 (+5.4\%)} &\cellcolor{lightmintgreen} \textbf{68.1 (+1.9\%)} \\
    \bottomrule
    \end{tabular}
    }
    \caption{\footnotesize \methodname{} outperforms CoOp for different CLIP backbones using labels generated by LLaMA for default $m,\kappa$=3.}
    \label{tab:different_backbone_small}
\end{wraptable}
All results in this paper are on the ViT-B/16 CLIP backbone. In this section we compare the performance of \methodname{}-LLaMA with CoOp-LLaMA across various other CLIP backbones. In Table~\ref{tab:different_backbone_small} we present the results of \methodname{} for a ResNet-50~\cite{He2015DeepRL}, ResNet-101, and a ViT-B/32 vision-encoder based CLIP model. We use the same configuration for each backbone. \methodname{} shows consistent improvement, achieving gains of $\mathbf{+4.1\%}$, $\mathbf{+1.8\%}$, and $\mathbf{+5.4\%}$ in average cACC for RN50, RN101 and ViT-B/32 respectively. These results highlight the effectiveness of our method over a diverse range of CLIP architectures. 

In Appendix \S~\ref{sec:appx_additional_results} we study different prompting strategies, effect of varying $\kappa$, impact of adaptive thresholding, variations across few-shot splits, and show qualitative results.

\vspace{-8pt}
\section{Conclusion}
\vspace{-5pt}

We addressed the challenge of Vocabulary-Free Fine-Grained Visual Recognition (VF-FGVR) by introducing \methodname{}, a method that leverages MLLMs to generate weakly supervised labels for a small set of training images, to efficiently fine-tune a downstream CLIP model. Our approach constructs a candidate label set for an image using generated labels of similar images, and performs label refinement for clean and noisy data differently. \methodname{} also proposes a label filtering strategy, effectively managing the open-ended and noisy nature of MLLM outputs. Experiments on 4 MLLMs show that \methodname{} significantly outperforms direct inference methods while dramatically reducing computational cost and inference time, setting a new benchmark for efficient and scalable VF-FGVR.

\section*{Impact Statement}
\label{sec:limitations}
\vspace{-2pt}

Our framework relies on MLLM-generated labels, a dependency that is becoming increasingly feasible with advancements in MLLM accessibility. We demonstrate strong performance across both proprietary (GPT-4o, GeminiPro) and open-source (LLaMA-11B, Qwen2-7B) models, showing robustness to MLLMs of varying capacities.  We see this work as a foundation for future research in leveraging MLLMs for fine-grained recognition, with no direct societal or ethical risks.


\clearpage
\section*{Appendix}
\setcounter{page}{1}
\setlength{\columnsep}{0.3125in}
\renewcommand{\thesection}{A\arabic{section}}
\renewcommand{\thetable}{A\arabic{table}}
\renewcommand{\thefigure}{A\arabic{figure}}

Our code will be made publicly available upon acceptance for further research and reproducibility. This supplementary material contains additional details that we could not include in the main paper due to space constraints, including the following information:

\begin{itemize}
    \item Further descriptions of datasets used
    \item Effect of label filtering on the size of label space
    \item Summary of notations and their descriptions
    \item The overall \methodname{} algorithm is presented in \S~\ref{sec appx algorithm}
    \item An analysis on an alternative way to obtain candidate sets is shown in \S~\ref{sec:appx_direct_candidates}
    \item Performance across different prompting strategies in shown in \S~\ref{app:prompting_strategies}
    \item The impact of varying $\kappa$, the number of nearest-neighbors considered is shown in \S~\ref{sec:appx_kappa}
    
    \item The effect of choice of threshold $\tau$ is studied in \S~\ref{sec:appx_threshold}
    \item Performance across variations in different few-shot splits is shown in  \S~\ref{app:different_seeds}
    \item Qualitative Results of \methodname{} against other baselines in \S~\ref{sec:qualitative_analysis}
    \item Discussion on other weaker open-source MLLMs is presented in \S~\ref{sec:appx_other_mllms}
    \item Expanded dataset-wise tables due to space constraints in the main paper are presented in \S~\ref{app:expanded_tables}
    \item Additional implementation details are given in \S~\ref{sec:appx_implementation}, where we discuss the hyperparameters of \methodname{}, present other details of open-source MLLMs, and show the prompts used to generate labels for different MLLMs 

\end{itemize}

\noindent\textbf{Description of Datasets Used.}
We show results on $5$ datasets with fine-grained labels -- Bird-200, Car-196, Dog-120, Flower-102, Pet-37. In Table~\ref{tab:datasetsummary}, we show the number of images used for training and the size of the test set.  

\begin{table}[H]
    \centering
    \scalebox{0.78}{
    \begin{tabular}{cccccc}
    \toprule
    &\textbf{Bird-200} &\textbf{Car-196}& \textbf{Dog-120}& \textbf{Flower-102}& \textbf{Pet-37}\\
    \midrule
      Train Set   & $m \times 200$ & $m \times 196$ &$m \times 120$&$m \times 102$&$m \times 37$\\
      \midrule
      Test Set   &5794 &8041&8550&6149&3669\\
    \bottomrule
    \end{tabular}
    }
    \caption{Train and test set sizes of the datasets used in this paper. The number of shots is denoted by $m$, with $m=3$ used as the default in our experiments unless otherwise specified.}
    \label{tab:datasetsummary}
\end{table}

\noindent\textbf{Effect of Label Filtering on the Size of Label Space.}
As described in \S~\ref{sec:ablation_study}, label filtering is crucial to obtain good VF-FGVR performance. In Table~\ref{tab:label_filtering}, we present the number of classes in the final classification label spaces that each method operates in. The first row indicates the size of the ground-truth label space. We observe that our label filtering mechanism is essential to combat the open-endedness of MLLM labels. 
\begin{table}
\scalebox{0.75}{
\centering
    \begin{tabular}{l|ccccc}
    \toprule
     \textbf{Method}    & \multicolumn{5}{c}{\textbf{Average}}\\
     \cmidrule{2-6}
         & Bird-200&Car-196&Dog-120&Flower-102&Pet-37\\
        \midrule
         Ground Truths&\large 200&\large 196&\large 120&\large 102&\large 37\\
         MLLM Labels &\large 412&\large 562&\large 169&\large 183&\large 63\\
         FineR &\large 202&\large 286&\large 97&\large 112&\large 44 \\
         NeaR-LLaMA &\large 239&\large 305&\large 129&\large 119&\large 45 \\
         \bottomrule
    \end{tabular}
    }
    \caption{Label filtering is effective in reducing the size of MLLM generated label to manageable levels.}
    \label{tab:label_filtering}
\end{table}

\noindent\textbf{Summary of Notations.} We represent elements of a set by a subscript and a vector component by a superscript. For instance $y_i\in\{0,1\}^{k}$ denotes the one-hot vector encoding of the class label of the $i$-th image, $i\in[n]$, and $y_{i}^{j}$ is the $j$-th component of this encoding. Specifically $y_{i}^{j}=1$ if the $i$-th image is assigned the $j$-th label in the label set $\mathcal{C}$ and $0$ otherwise. A summary of notations is given in Table~\ref{tab:notations}.

\begin{table}
\footnotesize
    \centering
    \scalebox{0.93}{
    \begin{tabular}{c|p{0.6\linewidth}}
    \toprule
    \textbf{Notation}&\textbf{Description}\\
    \midrule
    $x_i$ & $i$-th unlabeled training image\\
    $X$ & set of $n$ training images $\{x_1,x_2,\dots,x_n\}$\\
    $m$ & Number of shots of images for each class belonging to an unknown ground-truth class name set\\
    $l_i=L(x_i,p)$ & The class label generated for $x_i$ by an MLLM $L$ with a input prompt $p$\\
    $\mathcal{D}=\{(x_i,l_i)\}_{i=1}^{n}$ &The dataset generated by an MLLM $L$ consisting of image-class name pairs\\
    $\mathcal{C}=\bigcup_{i=1}^n l_i$ & Label space generated by the MLLM\\
    $k=\abs{\mathcal{C}}$& The $k$ lexicographically ordered class names $\mathcal{C}=\{c_1,c_2,\dots,c_k\}$\\
    $y_i\in\{0,1\}^{k}$ & One-hot encoding of the label $l_i$ of $x_i$\\
    $\mathcal{I}$ & Image encoder of pre-trained CLIP\\
    $\mathcal{T}$ & Text encoder of pre-trained CLIP\\
    $\theta$ & Learnable prompt vectors added to the input embeddings of a class name\\
    $f_{\theta}(x)\in\Delta^{k}$ & $k$-dimensional probability vector of the prompted CLIP model's class predictions for image $x$, i.e $f_{\theta}(x)^{j}\geq 0$ and $\sum\limits_{j=1}^{k} f_{\theta}(x)^{j}=1$\\
    $S_{i}$ & Candidate set created by gathering MLLM generated labels of nearest-neighbors of $x_i$\\
    $\kappa$ & Number of nearest-neighbors\\
    $\mathcal{D}_s = \{(x_i, l_i, S_i)\}_{i=1}^n$ & Augmented dataset containing the generated label $l_i$ and constructed candidate set $S_i$\\
    $L_{ce}(f_{\theta}(x_i),l_i)$ & The cross-entropy loss of CLIP model predictions for image $i$ w.r.t $y_i$ given by $-\sum_{j=1}^k y_{i}^{j}log(f_{\theta}^{j}(x_i))$ (also written as $L(x_i,l_i)$)\\
    $GMM$ & A two-component Gaussian Mixture Model fit on loss values of all training samples for every epoch\\ 
    $w_i=\mathbb{P}_{GMM}(clean|x_i)$ & The posterior probability of image belonging to the "clean" component, i.e component with lower mean\\
    $\tau\in[0,1]$& Threshold used to partition data into clean and noisy sets\\
    $X_{cl}, X_{ns}$& Clean and noisy partitions of the training data based on $w_\geq \tau$\\
    $q_i\in\mathbb{R}^k$& Confidence of the candidates. We have that $q_i^{j}>0$ if $c_j\in S_i$, $q_i^j=0$ otherwise, and $\sum\limits_{j=1}^k q_{i}^j=1$\\
    $\bar{y}_{i}$& Refined label for image $i$ constructed based on whether $x_i$ is clean or noisy. For noisy images we rescale the label to have non-zero probabilities only for the candidate labels\\
    shrp($y$, $T$) and rsc($y$, $q$) & sharpen a distribution $y$ using temperature $T$; rescale a distribution $y$ based on candidate confidence $q$\\
    $L_{final}(\theta)$ & The cross entropy loss between model predictions $f_{\theta}(x)$ and refined labels $\bar{y}$\\
    $\mathcal{C}_{test}$ & Inference label space post filtering\\
        \bottomrule
    \end{tabular}
    }
    \caption{List of notations used in our paper, and their descriptions.}
    \label{tab:notations}
\end{table}

\section{NeaR Algorithm}
\label{sec appx algorithm}

\newcommand{\mycomment}[1]{\textcolor{blue!75}{\STATEx /* \textit{#1} */}}
\begin{algorithm}
\footnotesize
\caption{\methodname{} algorithm: Training}
\begin{algorithmic}[1]
\REQUIRE $m$-shot training images $X=\{x_i\}_{i=1}^n$; MLLM $L$; input prompt $p$; number of nearest neighbors $\kappa$; CLIP model predictions $f_{\theta}$ with learnable text prompts $\theta$; $num\_epochs$; $warm\_epochs$; learning-rates $\eta$, $\eta_{warm}$; Temperature $T$ 
\ENSURE Trained parameters $\hat{\theta}$ 

\STATEx\hspace{-\algorithmicindent}\textbf{Step-1: Labeling training images with an MLLM}
\STATE $D\leftarrow \{\}$
\mycomment{Label each image $x$ by prompting MLLM $L$ with prompt $p$}
\FOR{$i=1,2,\dots,n$}
\STATE $D\leftarrow D\cup\{(x_i,l_i:=L(x_i,p))\}$
\ENDFOR
\STATE $\mathcal{C}:=\bigcup_{i=1}^n l_i$
\STATE $k:=\abs{\mathcal{C}}$
\mycomment{WLOG we consider $\mathcal{C}:=\{c_1,c_2,\dots,c_k\}$ where $k:=\abs{\mathcal{C}}$ to be lexicographically ordered. Let $y_i$ be the one-hot encoding of label $l_i$}
\STATE $y_i\in \{0,1\}^k$ and $y_{i}^{j}:=1$ if $c_j=l_i$ and $0$ o/w
\STATEx\hspace{-\algorithmicindent}\textbf{Step-2: Candidate Set Construction}
\STATE $D_{s}\leftarrow \{\}$
\mycomment{Augment each image $x_i$ with a candidate set $S_i$ composed of labels of $\kappa$-nearest neighbors}
\FOR{$(x_i,l_i) \text{ in } D$}
\STATE $S_i\leftarrow \text{knn\_labels}(x_i, \kappa)$
\STATE $D_s\leftarrow D_s\cup \{(x_i,l_i,S_i)\}$
\ENDFOR
\mycomment{We initialize candidate confidence for all images $q_i$ uniformly}
\STATE $q_i\in\mathbb{R}^k$ and $q_i^{j}:= \frac{1}{\abs{S_i}}$ if $c_j\in S_i$ and $0$ o/w
\STATEx\hspace{-\algorithmicindent}Function \texttt{Partition\_data($D$, $f_{\theta}$, $\tau$):}
    \STATE\hspace{\algorithmicindent} $\mathcal{L}:=\{L(f_{\theta}(x_i),l_i)\}_{i=1}^n$
    \STATE\hspace{\algorithmicindent} $\mu_c, \sigma_c, \mu_n, \sigma_n \leftarrow fit\_GMM(\mathcal{L})$
    \STATE\hspace{\algorithmicindent} $W:=\{w_1,w_2,\dots,w_n\}$ where $w_i=\mathbb{P}_{GMM}(clean|x_i)$
    \STATE\hspace{\algorithmicindent} $X_{cl}:=\{x_i\in X \mid w_i\geq \tau\}$
    \STATE\hspace{\algorithmicindent} $X_{ns}:=X\setminus X_{cl}$
\STATEx\hspace{-\algorithmicindent}\texttt{return} $X_{cl}, X_{ns}, W$
\STATEx\hspace{-\algorithmicindent}\textbf{Step-3: Fine-tune prompts $\theta$ of a CLIP model}
\FOR{$t=1,2,\dots,num\_epochs$}
    \mycomment{During warmup, the prompts are tuned on the cross-entropy loss $L_{ce}$ using MLLM generated labels in $D$}
    \IF{$t\leq warm\_epochs$}
    \STATE $\theta_{t}\leftarrow \theta_{t-1}-\eta_{warm}\nabla L_{ce}(D,f_{\theta_{t-1}})$
    \ELSE
    \mycomment{Every epoch post warm-up, we partition the entire data into clean and noisy sets by fitting a GMM on cross-entropy loss $L$}
    \STATE $X_{cl},X_{ns},W\leftarrow$\texttt{Partition\_data}($D$, $f_{\theta_{t-1}}$, $\tau$)
\STATE $\bar{y}_i:=\text{shrp}\big(w_i \cdot y_i + (1 - w_i) \cdot f_{\theta_{t-1}}(x_i), T\big), \text{\quad if } x_i\in X_{cl}$
    \STATEx \hspace{\algorithmicindent}\hspace{0.65cm}$:=\text{rsc}\big(\text{shrp}(w_i \cdot q_i + (1 - w_i) \cdot f_{\theta_{t-1}}(x_i), T), q_i\big), \text{ o/w}$
    \mycomment{We update candidate confidence (for both clean and noisy samples) to be used in the next epoch. $1[q_i]$ is $1$ at non-zero indices and $0$ o/w}
    \STATE $q_i\leftarrow \text{rsc}(f_{\theta_{t-1}}(x_i), 1[q_i]) $
    \STATE $L_{final}(\theta_{t-1}):=\frac{-1}{n}\sum\limits_{i=1}^n\sum\limits_{j=1}^{k} \bar{y}_{i}^{j}\cdot log(f_{\theta_{t-1}}^{j}(x_i))$
    \STATE $\theta_{t}\leftarrow \theta_{t-1}-\eta\nabla L_{final}(\theta_{t-1})$
    \ENDIF
\ENDFOR
\STATE\textbf{Return:} $\hat{\theta}=\theta_{num_{epochs}}$
\end{algorithmic}
\label{mainalgo}
\end{algorithm}

We present the training algorithm of \methodname{} in Algorithm~\ref{mainalgo}. We present the three steps of our method \methodname{} as shown in Figure~\ref{fig:mainfigure}. In lines L1-L7 we generate possibly noisy labels from an MLLM. In lines L8-L12, we then generate a candidate set from $\kappa$ nearest-neighbors of each image. In lines L19-L27 we train prompt vectors $\theta$ by minimizing the cross-entropy loss between CLIP predictions and the refined labels. The label refinement in L24 effectively disambiguates the best label from the generated candidates. As training progresses, our estimate of the candidate labels $q_i$ gets better. The sharpening function is $\text{shrp}(y,T)^{i}=(y^{i})^{\frac{1}{T}}/\sum\limits_{j=1}^{k} (y^{j})^{\frac{1}{T}}$ and the rescale function is defined as $\text{rsc}(y, q)^{i} = (y\odot q)^{i}/\sum\limits_{j=1}^k (y\odot q)^{j}$.

\section{Additional Results}
\label{sec:appx_additional_results}
In this section we present the following results 
i) In \S~\ref{sec:appx_direct_candidates} we analyze an alternative approach to generating candidate sets and compare it with our proposed $\kappa$-NN based approach. ii) Different Prompting Strategies like text, visual and both are explored in \S~\ref{app:prompting_strategies}. iii) In \S~\ref{sec:appx_kappa} we study the impact of varying the number of nearest-neighbors $\kappa$ on the performance of \methodname{}; iv)  In \S~\ref{sec:appx_threshold} we study the impact of our adaptive thresholding. v) Performance across variations in different few-shot splits is studied in \S~\ref{app:different_seeds}. vi) Qualitative results are presented in \S~\ref{sec:qualitative_analysis}. vii) In S~\ref{sec:appx_other_mllms} we present results on two weaker open-source MLLMs -- LLaVA-1.5, and BLIP2;  viii) Expanded tables of Comparison of \methodname{} with JoAPR, PRODEN, on Imbalanced training data and across different backbones in \S~\ref{app:expanded_tables}.


\subsection{Analysis on Alternative Ways to Construct Candidate Sets}
\label{sec:appx_direct_candidates}
The labels generated by MLLMs can be noisy. To address this, we propose to construct a candidate set for each image by grouping class labels from the $\kappa$ nearest-neighbors of the image. In this section we study an alternative approach to candidate set generation, where we query the MLLM itself to generate a set of $\kappa$ labels directly for each image. We present the results of this approach in Table~\ref{tab:mllm_candidateset}, showing performance across three MLLMs: GPT-4o, GeminiPro, and LLaMA.

\begin{table*}
    \centering
    \scalebox{0.78}{
    \begin{tabular}{l|cccccccccc|cc}
    \toprule
    \textbf{Method} & \multicolumn{2}{c}{\textbf{Bird-200}} & \multicolumn{2}{c}{\textbf{Car-196}} & \multicolumn{2}{c}{\textbf{Dog-120}} & \multicolumn{2}{c}{\textbf{Flower-102}} & \multicolumn{2}{c}{\textbf{Pet-37}} & \multicolumn{2}{|c}{\textbf{Average}} \\
    \cmidrule{2-13}
         & cACC & sACC & cACC & sACC & cACC & sACC & cACC & sACC & cACC & sACC & cACC & sACC \\
        \midrule
        \methodname{}-GPT-4o-Direct&53.1&74.2&OOM&OOM&61.0&71.9&77.0&55.4&83.1&82.7&-&-\\
        \methodname{}-GPT-4o&55.8&75.6&57.0&60.0&61.6&74.4&80.6&52.1&82.9&84.0&67.6&69.2\\
        \midrule
        \methodname{}-GeminiPro-Direct&52.7&73.8&OOM&OOM&70.5&58.3&73.9&46.9&81.3&81.7&-&-\\
        NeaR-GeminiPro&55.9&76.0&54.9&61.1&64.7&75.4&77.9&53.2&79.4&80.8&66.6&69.3\\
        \midrule
        \methodname{}-LLaMA-Direct &43.8&68.2&44.7&56.0&56.4&70.6&70.9&57.7&83.2&87.5&59.8&68.0\\
        \rowcolor{lightmintgreen}NeaR-LLaMA&51.0&70.2&52.6&60.9&59.2&70.2&78.6&61.7&83.5&86.2&\textbf{65.0\textbf{ (+5.2\%)}}&\textbf{69.8 (+1.8\%)}\\
         \bottomrule
    \end{tabular}
    }
    \caption{Evaluation of NeaR-MLLM under different candidate set generation methods. We compare our $\kappa$-nn-based candidate set against directly querying the MLLM for a candidate set, referred to as NeaR-MLLM-Direct. For the Car-196 dataset, both GPT-4o and GeminiPro encounter Out-of-Memory (OOM) errors due to the larger label space. For NeaR-LLaMA, our $\kappa$-nn-based approach outperforms the direct approach by an average margin of $+5.8\%$ in cACC while being more computationally efficient.}
    \label{tab:mllm_candidateset}
    \vspace{-5pt}
\end{table*}

\begin{table*}
    \centering
    \scalebox{0.78}{
    \begin{tabular}{l|cccccccccc|cc}
    \toprule
    \textbf{Method} & \multicolumn{2}{c}{\textbf{Bird-200}} & \multicolumn{2}{c}{\textbf{Car-196}} & \multicolumn{2}{c}{\textbf{Dog-120}} & \multicolumn{2}{c}{\textbf{Flower-102}} & \multicolumn{2}{c}{\textbf{Pet-37}} & \multicolumn{2}{|c}{\textbf{Average}} \\
    \cmidrule{2-13}
         & cACC & sACC & cACC & sACC & cACC & sACC & cACC & sACC & cACC & sACC & cACC & sACC \\
         \midrule
         \multicolumn{12}{c}{Text Prompting} \\
         \midrule
        CoOp-LLaMA&49.2&68.7&45.5&60.7&58.4&68.4&75.9&59.8&74.4&79.2&60.7&67.4\\
        \rowcolor{lightmintgreen}NeaR-LLaMA&51.0&70.2&52.6&60.9&59.2&70.2&78.6&61.7&83.5&86.2&$\mathbf{65.0{ \textbf{ (+4.3\%)}}}$&$\mathbf{69.8 \textbf{ (+2.4\%)}}$\\
        \midrule
        \multicolumn{12}{c}{Visual Prompting} \\
        \midrule
        VPT-LLaMA &48.9&69.5&45.3&61.3&60.3&70.2&78.1&61.7&73.2&82.2&61.1&69.0\\
        \rowcolor{lightmintgreen}NeaR-VPT-LLaMA&50.2&68.5&45.5&60.1&59.4&71.4&78.3&61.3&81.1&84.3&$\mathbf{62.9 \textbf{ (+1.8\%)}}$&$\mathbf{69.1} \textbf{ (+0.1\%)}$\\
        \midrule
        \multicolumn{12}{c}{Multimodal Prompting} \\
        \midrule
        IVLP-LLaMA&48.9&69.5&45.3&61.3&60.3&70.2&78.1&61.7&73.2&82.2&61.1&69.0\\
        \rowcolor{lightmintgreen}NeaR-IVLP-LLaMA&50.8&70.1&52.5&61.0&58.6&69.9&80.3&62.1&83.7&86.4&$\mathbf{65.2} \textbf{ (+4.1\%)}$&$\mathbf{69.9 \textbf{ (+0.9\%)}}$\\
        \bottomrule
    \end{tabular}
    }
    \caption{Evaluation of \methodname{} under different prompting strategies. In addition to text-based prompting, as shown in Tab~\ref{tab:maintable}, we present results on Visual Prompting method VPT~\cite{Jia2022VisualPT} and Multimodal Prompting method IVLP~\cite{ivlp}. We outperform the baselines by $+1.8\%$ and $+4.1\%$ in cACC respectively.}
    \label{tab:prompting_strategies}
    \vspace{-5pt}
\end{table*}

\begin{table*}
    \centering
    \scalebox{0.78}{
    \begin{tabular}{l|cccccccccc|cc}
    \toprule
    \textbf{Method} & \multicolumn{2}{c}{\textbf{Bird-200}} & \multicolumn{2}{c}{\textbf{Car-196}} & \multicolumn{2}{c}{\textbf{Dog-120}} & \multicolumn{2}{c}{\textbf{Flower-102}} & \multicolumn{2}{c}{\textbf{Pet-37}} & \multicolumn{2}{|c}{\textbf{Average}} \\
    \cmidrule{2-13}
         & cACC & sACC & cACC & sACC & cACC & sACC & cACC & sACC & cACC & sACC & cACC & sACC \\
         \midrule
        NeaR-LLaMA ($\tau$=0.5)&51.7&70.5&53.3&60.5&59.0&68.6&78.1&61.7&82.2&85.9&64.8&69.5\\
        \rowcolor{lightmintgreen}NeaR-LLaMA&51.0&70.2&52.6&60.9&59.2&70.2&78.6&61.7&83.5&86.2&\textbf{65.0 (+0.2\%)}&\textbf{69.8 (+0.3\%)}\\
        \midrule
        NeaR-GPT-4o ($\tau$=0.5) &54.7&74.5&57.9&59.7&62.1&74.6&79.6&52.1&83.0&83.8&67.4&68.9\\
        \rowcolor{lightmintgreen}NeaR-GPT-4o&55.8&75.6&57.0&60.0&61.6&74.4&80.6&52.1&82.9&84.0&\textbf{67.6 (+0.2\%)}&\textbf{69.2 (+0.3\%)}\\
        \midrule
        NeaR-GeminiPro ($\tau$=0.5) &52.8&73.5&53.7&61.1&64.8&75.2&77.6&53.3&77.7&81.0&65.3&68.8\\
        \rowcolor{lightmintgreen}NeaR-GeminiPro&55.9&76.0&54.9&61.1&64.7&75.4&77.9&53.2&79.4&80.8&\textbf{66.6 
 (+1.3\%)}&\textbf{69.3 (+0.5\%)}\\
        \midrule
        NeaR-Qwen2 ($\tau$=0.5) &34.3&65.0&57.0&64.0&55.2&71.3&74.0&61.4&74.1&76.4&58.9&67.6\\
        \rowcolor{lightmintgreen}NeaR-Qwen2&35.5&65.6&58.0&64.0&56.6&71.7&75.8&62.5&73.8&76.4&\textbf{60.0 (+1.1\%)}&\textbf{68.0 (+0.4\%)}\\
         \bottomrule
    \end{tabular}
    }
    \caption{Evaluation of our dynamic threshold $\tau$ across different MLLMs compared to a static threshold $\tau=0.5$. The use of a dynamic threshold shows consistent improvements across all MLLMs, with gains in NeaR-GeminiPro and NeaR-Qwen2, achieving increases of $1.6\%$ and $1.1\%$ in average cACC, respectively, and a minor gain of $0.2\%$ in other cases. These results support the design choice to avoid the hyperparameter $\tau$, which can vary slightly across MLLMs.}
    \label{tab:threshold_expts}
    \vspace{-5pt}
\end{table*} 

\subsection{Different Prompting Strategies.}
\label{app:prompting_strategies}
 We analyze the impact of our proposed approach using various prompting strategies, as presented in Table~\ref{tab:prompting_strategies}. We consider three distinct prompting methods involving fine-tuning across different modalities: (1) For text-only prompting, we use CoOp~\cite{coop}; (2) For image-only prompting, we employ VPT~\cite{Jia2022VisualPT}; and (3) For both text and image prompting, we adopt hierarchical prompts introduced at different text and image layers~\cite{ivlp} as the backbone. Our method demonstrates strong performance across all prompting strategies, achieving cACC improvements of $4.2\%$, $1.9\%$, and $4.1\%$, respectively. This clearly demonstrates the effectiveness of our method across different fine-tuning methods.

\subsection{Impact of Varying No. of Nearest-Neighbors $\kappa$}
\label{sec:appx_kappa}
\begin{figure}
    \centering
    \includegraphics[width=0.75\linewidth]{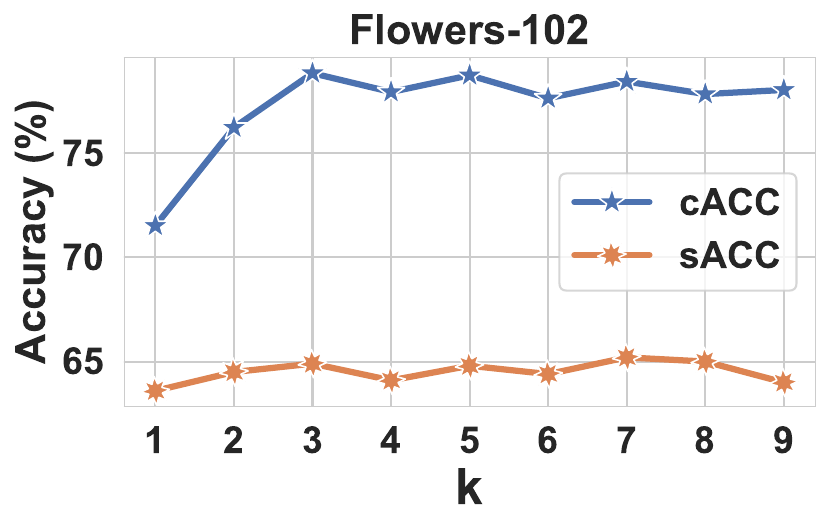}
    \caption{Effect of varying $\kappa$ (1 to 9) on the performance of NeaR-LLaMA for the 9-shot Flowers-102 dataset. The results show an upward trend in cACC as $\kappa$ increases from 1 to 3, reflecting an increased likelihood of semantically closer labels. However, for $\kappa \geq 3$, the performance plateaus or slightly decreases due to a noisier candidate set, validating our choice of $\kappa=3$.}
    \label{fig:k_ablation}
\end{figure}

We leverage similarity information to build a candidate set for each image by augmenting its label with the labels of its $\kappa$ nearest-neighbors. In this section we study the effect of varying $\kappa$ from $1$ to $9$ on the performance of \methodname{}-LLaMA. We perform this experiment for $9$-shot data from the Flowers-102 dataset, to ensure that higher values of $\kappa$ give meaningful results. The results in Figure~\ref{fig:k_ablation} show that \methodname{} performs well across a large range of $\kappa$ values, and justifies our choice of $\kappa=3$. Note that setting $\kappa=1$ is not the same as CoOp-LLaMA, but is the result of \methodname{} with $q_i=y_i$. The results also highlight two competing factors that influence the performance of \methodname{} as $\kappa$ varies:
\begin{itemize}
    \item Improved Label Quality with Larger Candidate Sets -- A larger candidate set is more likely to involve a semantically closer label. This is reflected in the upward trend of cACC from $\kappa=1$ to $\kappa=3$.
    \item Increased Noise with Larger Candidate Sets -- For higher values of $\kappa$, while the likelihood of including better labels in the candidate set increases, it is offset by the addition of irrelevant labels. A noisier candidate set makes it harder for the algorithm to disambiguate the best label in the candidate set. This leads to a plateau or even slight decrease in cACC for $\kappa>3$.    
\end{itemize}

Our proposed approach constructs candidate sets from single labels assigned to each image, while the direct candidate set method generates a set of labels for each image. A notable drawback of the direct approach is the increase in size of the final label space, which may become prohibitively large as each image contributes to $\kappa-1$ new labels in the worst case. The direct method thus incurs a larger memory footprint and requires longer training times due to the larger label space. For both GPT-4o and GeminiPro, the direct approach termed \methodname{}-MLLM-Direct encounters an out-of-memory (OOM) error on the Car-196 dataset. Furthermore, for LLaMA, our $\kappa$-nn based approach outperforms the direct approach by a substantial margin, achieving a $\mathbf{+5.8\%}$ higher cACC, while being more efficient.



\subsection{Effect of Choice of Threshold $\tau$}
\label{sec:appx_threshold}
To address the noisy nature of MLLM generated labels, our method \methodname{} separates samples into clean and noisy sets using a threshold $\tau$ based on clean posterior probability $w_i$ of a GMM fitted on loss values. Instead of using a fixed threshold, we make $\tau$ adaptive by setting it to the mean posterior probability, $\tau = \frac{1}{n} \sum\limits_{i=1}^n w_i$, allowing dynamic estimation of label noise at every training epoch. We study the effects of using a fixed threshold of $\tau=0.5$ for \methodname{}-LLaMA, NeaR-GPT-4o, NeaR-GeminiPro and NeaR-Qwen2 in Table~\ref{tab:threshold_expts}. We observe that for NeaR-GeminiPro and NeaR-Qwen2, we have a performance gain of $+1.3\%$ and $+1.1\%$ in average cACC, while a relatively lower performance gain of $+0.2\%$ in NeaR-LLaMA and NeaR-GPT-4o. These results show that our adaptive thresholding performs better than a static threshold across a variety of MLLM choices, thus eliminating the need for tuning the hyperparameter $\tau$.

\subsection{Performance across variations in different few-shot splits}
\label{app:different_seeds}
\begin{figure}
    \centering
    \includegraphics[width=0.99\linewidth]{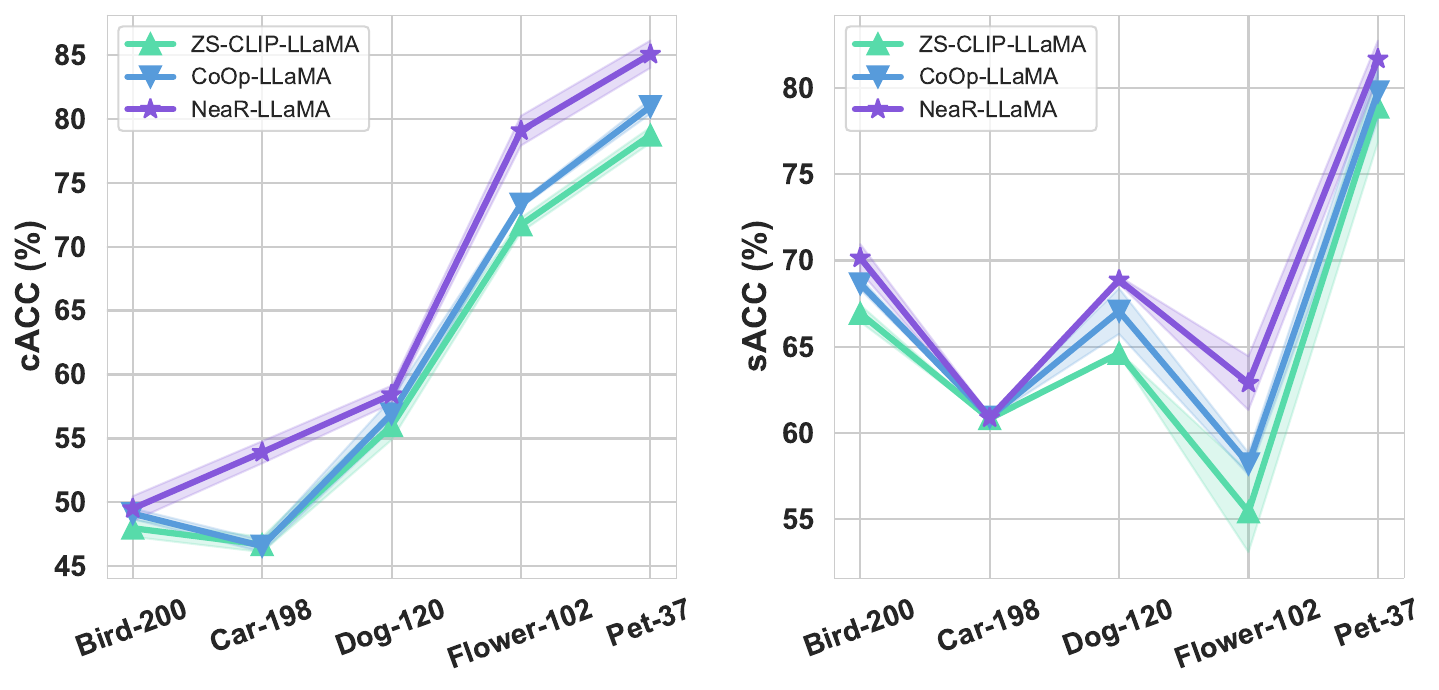}
    \caption{We report cACC and sACC under the effect of random sampling of training images across five datasets. The plot demonstrates minimal variance across datasets, highlighting the robustness of \methodname{} to variations in data selection.}
    \label{fig:llamaseeds}
\end{figure}

We have conducted extensive experiments across three random seeds to evaluate the robustness and consistency of our approach as shown in Fig~\ref{fig:llamaseeds}. Specifically, for each seed, we sampled a unique set of $m$ images per class, ensuring diversity in the training data distribution across different runs. The results consistently demonstrate that \methodname{} outperforms the baseline approaches across various independent samplings, with minimum variance across datasets, indicating its stability and generalizability. This robustness across different random seeds highlights the effectiveness of our approach in handling variations in training data selection, further strengthening its practical applicability in real-world scenarios.

\subsection{Qualitative Results of \methodname{} against other baselines.}
\label{sec:qualitative_analysis}
We visualize a selection of inference images and analyze their predictions across different models. Specifically, we compare the predictions of \methodname{} against those obtained from directly querying the MLLM, as well as the outputs of ZS-CLIP-LLaMA and CoOp-LLaMA. On the Bird-200 dataset(row 1), \methodname{} successfully predicts the exact ground-truth label in cases where the MLLM itself fails, demonstrating the effectiveness of our candidate set in guiding classification. Additionally, even in instances where \methodname{}'s predictions do not match the ground truth, the errors are less severe compared to those of other baselines. For example, for Audi S5 Convertible and Old English Sheepdog, \methodname{} predicts Audi S5 and English Sheepdog, respectively. While these predictions deviate slightly from the exact ground-truth labels (Audi S5 Coupe and Old English Sheepdog), they remain semantically closer and more reasonable than the mistakes made by other methods.

\subsection{Results on Other Open-Source MLLMs}
\label{sec:appx_other_mllms}
In order to study the impact of \methodname{} on other open-source MLLMs, we query two weaker open-source MLLMs, LLaVA-1.5~\cite{liu2023improvedllava}and BLIP2~\cite{li2023blip}, to generate labels for our datasets. In the context of addressing the VF-FGVR problem, we observe that these models produce generic labels that lack fine-grained detail. For instance, in the Bird-200 dataset, images from various fine-grained classes such as American Goldfinch, Tropical Kingbird, Blue-headed Vireo, Yellow-throated Vireo, Blue-winged Warbler, Canada Warbler, Cape-May Warbler, and Palm Warbler were all labeled simply as \texttt{`Bird'} by LLaVA. This lack of specificity results in a low cACC of $9.8\%$ for CoOp-LLaVA and $4.7\%$ for \methodname{}-LLaVA. This trend is also observed with BLIP2. The inability of these MLLMs to generate diverse fine-grained labels makes them a poor choice to solve the VF-FGVR task.

\subsection{Expanded Tables}
\label{app:expanded_tables}
Due to the space constraints in the main paper, we present the detailed dataset-wise results on Comparison of \methodname{} with contemporary noisy label learning method and partial label learning method in ~\ref{tab:compare_with_joapr_and_proden}, Comparison of \methodname{} with other baselines under a long-tail class-imbalanced setting in Tab~\ref{tab:long_tail} and Comparison of \methodname{} with baselines across different backbones in shown in Tab~\ref{tab:different_backbone}. 
\begin{table*}[ht]
    \centering
    \scalebox{0.8}{
    \begin{tabular}{l|cccccccccc|cc}
    \toprule
    \textbf{Method} & \multicolumn{2}{c}{\textbf{Bird-200}} & \multicolumn{2}{c}{\textbf{Car-196}} & \multicolumn{2}{c}{\textbf{Dog-120}} & \multicolumn{2}{c}{\textbf{Flower-102}} & \multicolumn{2}{c}{\textbf{Pet-37}} & \multicolumn{2}{|c}{\textbf{Average}} \\
    \cmidrule{2-13}
         & cACC & sACC & cACC & sACC & cACC & sACC & cACC & sACC & cACC & sACC & cACC & sACC \\
         \toprule
         JoAPR-LLaMA &49.2&70.0&42.8&60.6&59.5&70.6&76.7&60.1&73.9&83.3&60.4&68.9\\
        PRODEN &48.3&67.6&45.9&60.6&57.9&67.0&75.2&59.0&75.8&78.5&60.6&66.6\\
        \rowcolor{lightmintgreen}NeaR-LLaMA     & 51.1 & 70.2 & 52.5 & 60.8 & 59.2 & 70.2 & 78.6 & 61.7 & 83.4 & 86.1 & \textbf{64.9 {(+4.3\%)}} & \textbf{69.8 {(+3.2\%)}}\\
         \bottomrule
    \end{tabular}
    }
    \caption{Comparison of NeaR with a contemporary noisy label learning method, JoAPR~\cite{guo2024joapr}, for CLIP using LLaMA-generated labels. NeaR outperforms JoAPR with an average improvement of $+4.6\%$ in cACC and $+0.9\%$ in sACC. These results indicate that directly applying LNL methods is insufficient to handle the challenges of noisy MLLM outputs. By incorporating better label refinement using candidate set, and by performing label filtering, \methodname{} provides a robust solution to the VF-FGVR problem. We also compare our method against PRODEN\cite{provablyPLL}. We significantly outperform PRODEN on both cACC and sACC.}
    \label{tab:compare_with_joapr_and_proden}
    \vspace{-5pt}
\end{table*}

\begin{table*}[ht]
    \centering
    \scalebox{0.82}{
    \begin{tabular}{l|cccccccccc|cc}
    \toprule
    \textbf{Method} & \multicolumn{2}{c}{\textbf{Bird-200}} & \multicolumn{2}{c}{\textbf{Car-196}} & \multicolumn{2}{c}{\textbf{Dog-120}} & \multicolumn{2}{c}{\textbf{Flower-102}} & \multicolumn{2}{c}{\textbf{Pet-37}} & \multicolumn{2}{|c}{\textbf{Average}} \\
    \cmidrule{2-13}
         & cACC & sACC & cACC & sACC & cACC & sACC & cACC & sACC & cACC & sACC & cACC & sACC \\
         \midrule
        FineR&46.2&66.6&48.5&62.9&42.9&61.4&58.5&48.2&63.4&67.0&51.9&61.2\\
        ZS-CLIP-LLaMA &48.9&67.0&46.9&60.3&55.9&64.5&71.4&58.5&75.5&72.2&59.7&64.5\\
        CoOp-LLaMA&47.9&69.9&45.6&60.6&54.2&67.8&74.0&60.2&78.0&74.0&60.0&66.5\\
        \rowcolor{lightmintgreen}NeaR-LLaMA     & 50.9 & 69.9 & 52.6 & 60.4 & 60.2 & 71.2 & 80.3 & 63.8 & 84.6 & 86.2 & \textbf{65.7 {\textbf{(+5.7\%)}}} & \textbf{70.3 {\textbf{(+3.8\%)}}}\\
         \bottomrule
    \end{tabular}
    }
    \caption{Performance comparison of NeaR-LLaMA with other baselines under long-tail class distribution. Both NeaR and CoOp retain performance on imbalanced data compared to balanced sampling. NeaR outperforms CoOp by $+5.7\%$ in cACC.}
    \label{tab:long_tail}
\end{table*}

\begin{table*}
    \centering
    \scalebox{0.84}{
    \begin{tabular}{l|cccccccccc|cc}
    \toprule
    \textbf{Method} & \multicolumn{2}{c}{\textbf{Bird-200}} & \multicolumn{2}{c}{\textbf{Car-196}} & \multicolumn{2}{c}{\textbf{Dog-120}} & \multicolumn{2}{c}{\textbf{Flower-102}} & \multicolumn{2}{c}{\textbf{Pet-37}} & \multicolumn{2}{|c}{\textbf{Average}} \\
    \cmidrule{2-13}
         & cACC & sACC & cACC & sACC & cACC & sACC & cACC & sACC & cACC & sACC & cACC & sACC \\
         \midrule
         \multicolumn{12}{c}{RN50} \\
         \midrule
        CoOp-LLaMA &13.9&44.1&8.6&46.6&18.2&47.9&15.6&31.7&36.7&54.7&18.6&45.0\\
        \rowcolor{lightmintgreen}NeaR-LLaMA&17.3&50.5&9.8&48.9&19.9&49.9&20.5&36.6&46.1&63.3&\textbf{22.7 (+4.1\%)}&\textbf{49.8 (+4.8\%)}\\
        \midrule
        \multicolumn{12}{c}{RN101} \\
        \midrule
        CoOp-LLaMA &17.3&45.5&9.6&47.4&20.8&45.4&17.5&34.7&44.0&55.2&21.8&45.6\\
        \rowcolor{lightmintgreen}NeaR-LLaMA&18.6&49.6&11.4&50.8&22.4&51.8&18.9&36.1&47.0&63.9&\textbf{23.6 (+1.8\%)}&\textbf{50.5 (+4.9\%)}\\
        \midrule
        \multicolumn{12}{c}{ViT-B/16} \\
        \midrule
         CoOp-LLaMA&49.2&68.7&45.5&60.7&58.4&68.4&75.9&59.8&74.4&79.2&60.7&67.4\\
        \rowcolor{lightmintgreen}NeaR-LLaMA&51.0&70.2&52.6&60.9&59.2&70.2&78.6&61.7&83.5&86.2&\textbf{65.0 (+4.3\%)}&\textbf{69.8 (+2.4\%)}\\
        \midrule
        \multicolumn{12}{c}{ViT-B/32} \\
        \midrule
        CoOp-LLaMA &45.0&56.3&39.3&60.5&51.9&66.2&69.8&59.1&72.7&79.8&55.7&66.2\\
        \rowcolor{lightmintgreen}NeaR-LLaMA&48.8&68.4&47.8&60.0&56.4&68.9&75.0&61.3&77.3&82.0&\textbf{61.1 (+5.4\%)}&\textbf{68.1 \textbf{(+1.9\%)}}\\
        \bottomrule
    \end{tabular}
    }
    \caption{Performance comparison of NeaR-LLaMA with CoOp-LLaMA across different CLIP backbones, including ResNet-50 (RN50), ResNet-101 (RN101), and ViT-B/32. For completeness, results are also provided for the default backbone, ViT-B/16. NeaR consistently outperforms CoOp-LLaMA, achieving gains of $+4.1\%$ and $+4.8\%$ in cACC and sACC for RN50, $+1.8\%$ and $+4.9\%$ in cACC and sACC for RN101, and $+5.4\%$ and $+1.9\%$ in cACC and sACC for ViT-B/32.}
    \label{tab:different_backbone}
    \vspace{-5pt}
\end{table*}

\begin{figure*}
    \centering
    \includegraphics[width=0.99\linewidth]{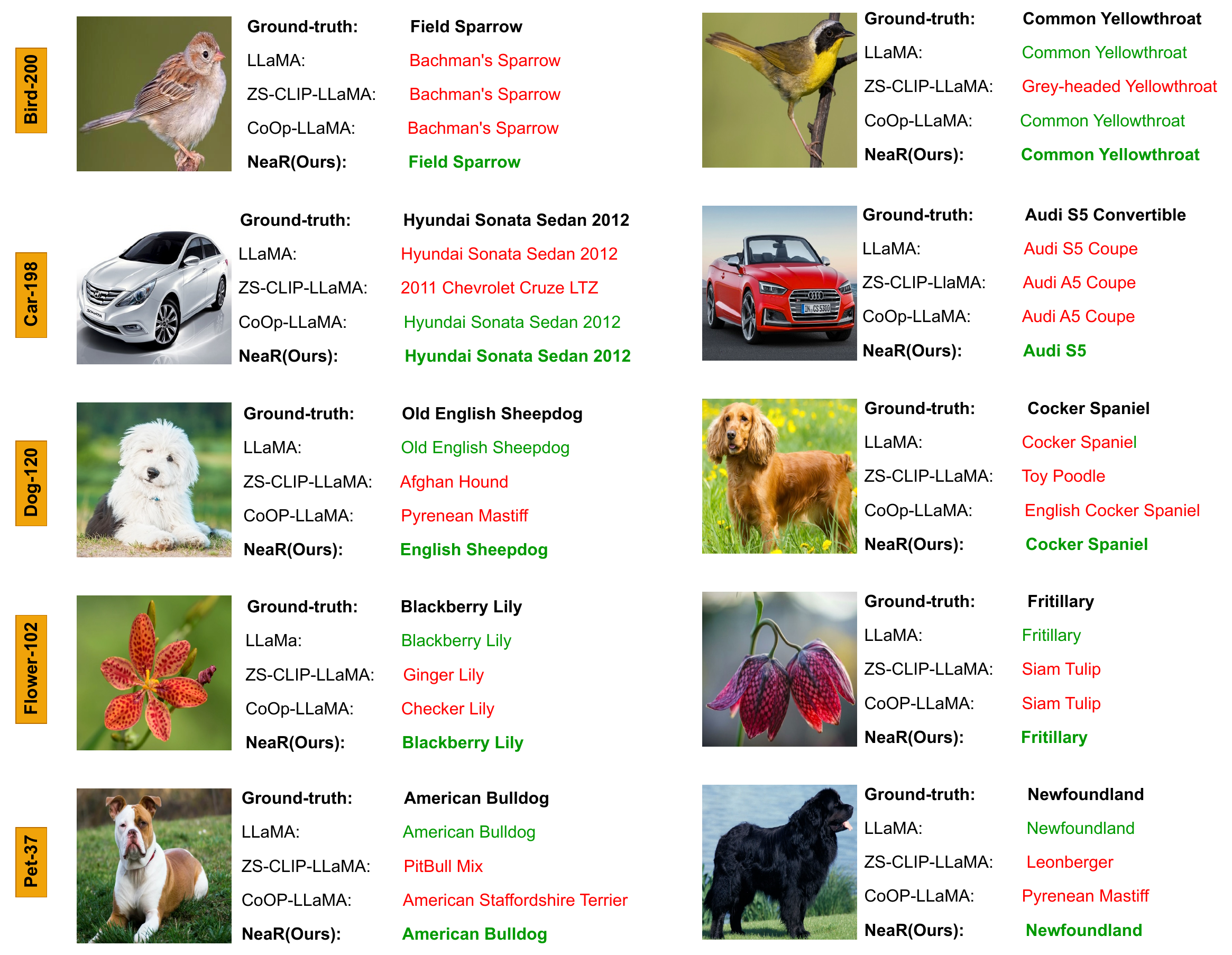}
    \caption{We present qualitative results across five benchmark datasets, comparing predictions from LLaMA, ZS-CLIP-LLaMA, CoOp-LLaMA, and \methodname{}.}
    \label{fig:Qualitativeresults}
\end{figure*}


\vspace{-2pt}
\section{Implementation Details}
\label{sec:appx_implementation}
Apart from the implementation details mentioned in the main paper, we present a few more details. We use a temperature of 2 in the sharpening function. Our batch size is $32$. We use the SGD optimizer with a learning rate of $0.002$, and use both constant learning rate scheduler and cosine annealing scheduler sequentially. The training hyperparameters are the same for CoOp and NeaR. We sample an equal batch of clean and noisy samples during every epoch. We run all our experiments on a single Nvidia Tesla V100-32GB GPU with an Nvidia driver version of 525.85.12. We use PyTorch 2.4.0 and CUDA 12.0. The default value of number of nearest-neighbors $\kappa$ is $3$, and the number of shots $m$ is $3$. We use the few-shot splits provided by FineR~\cite{finer}.

\noindent\textbf{Open-Source MLLM details.}
We utilize the publicly available meta-llama/Llama-3.2-11B-Vision-Instruct model and Qwen/Qwen2-VL-2B-Instruct model from HuggingFace. We observe that instruction tuned MLLMs generate better labels compared to base models. We perform inference using the HuggingFace \texttt{transformers} library~\cite{huggingface_transformers}. 

\noindent\textbf{Prompts used to generate labels from MLLMs.} In Table~\ref{tab:prompts}, we describe the prompts used to obtain the labels for both proprietary and open-source MLLMs. We give different prompts for different datasets. As part of future work, we would like to explore if different prompting strategies can give better labels.

\begin{table*}
    \centering
    \begin{tabular}{c|c}
    \toprule
    \textbf{MLLM} & \textbf{Prompt Structure}\\
    \midrule
        & \texttt{``You are a multimodal  AI trained to provide the best } \\
     GPT-4o, GeminiPro  & \texttt{fine-grained class label for a given <dataset> image.}\\
       &\texttt{Provide the best fine-grained class label for the given}\\
      &\texttt{<dataset> image. Do not return anything else.'', <img>}  \\
      \midrule
        & \texttt{``Give me a fine-grained label for this <dataset>.}\\
         LLaMA, Qwen & \texttt{For example, <samplelabel>.} \\
        &\texttt{Just print the label and nothing else.'', <img>}\\
    \bottomrule
    \end{tabular}
    \caption{A summary of prompts used for querying MLLM models used in this paper. In these prompts, $\texttt{dataset}\in \{bird, car, dog, flower, pet\}$. \texttt{img} refers to the image being queried for fine-grained class label. \texttt{Samplelabel} for \textit{bird} is \texttt{Black Throated Sunbird}, \texttt{samplelabel} for \textit{car} is \texttt{2012 BMW M3 coupe}, etc. We observe that open-source models like LLaMA require extra supervision in terms of sample labels for better performance.}
    \label{tab:prompts}
\end{table*}

\end{document}